\documentclass[journal,transmag]{IEEEtran}

\usepackage{xcolor,soul,framed} %,caption

\colorlet{shadecolor}{yellow}
\usepackage[pdftex]{graphicx}
\graphicspath{{../pdf/}{../jpeg/}}
\DeclareGraphicsExtensions{.pdf,.jpeg,.png}

\usepackage{adjustbox}
\usepackage{mdwmath}
\usepackage{eqparbox}
\usepackage{url}
\usepackage[noadjust]{cite}

\usepackage{fancyhdr}
\usepackage{xcolor}
\usepackage{eso-pic}
\usepackage{forloop}
\usepackage{algorithm}
\usepackage{algpseudocode}
\usepackage{microtype}
\usepackage{booktabs}
\usepackage{amsmath}
\usepackage{amssymb}
\usepackage{mathtools}
\usepackage{amsthm}
\usepackage{amsfonts}
\usepackage{colortbl}
\usepackage{threeparttable}
\usepackage{array}
\usepackage{float}
\usepackage{placeins}
\usepackage{subcaption}
\usepackage{booktabs}
\usepackage{diagbox} %% Array diagonal box
\usepackage{multirow}
\usepackage{multicol}
\usepackage{graphicx}
\definecolor{lightblue}{rgb}{0.9, 0.95, 1.0}
\definecolor{mygreen}{rgb}{0.01, 0.5, 0.01}
\definecolor{myred}{rgb}{0.8, 0.01, 0.01}
\usepackage{soul}
\usepackage{longtable}
\usepackage[pagebackref,breaklinks,colorlinks,allcolors=purple]{hyperref}
\def\BibTeX{{\rm B\kern-.05em{\sc i\kern-.025em b}\kern-.08em
    T\kern-.1667em\lower.7ex\hbox{E}\kern-.125emX}}

\usepackage{mathrsfs}

\usepackage[switch]{lineno}

\usepackage[moderate,tracking=normal,paragraphs=normal]{savetrees} % paragraphs=normal,tracking=normal,

\begin{document}
\bstctlcite{IEEEexample:BSTcontrol}
    \title{DyCAF-Net: Dynamic Class-Aware Fusion Network}

% \author{Anonymous Authors}

\author{
\IEEEauthorblockN{
Md Abrar Jahin\IEEEauthorrefmark{1},
Shahriar Soudeep\IEEEauthorrefmark{2},
M. F. Mridha\IEEEauthorrefmark{2},
Nafiz Fahad\IEEEauthorrefmark{3}, and
Md. Jakir Hossen\IEEEauthorrefmark{3}\\
}
\IEEEauthorblockA{\IEEEauthorrefmark{1}\textit{\small Thomas Lord Department of Computer Science, University of Southern California, Los Angeles, CA 90089, USA}}\\
\IEEEauthorblockA{\IEEEauthorrefmark{2}\textit{\small Department of Computer Science, American International University-Bangladesh, Dhaka 1229, Bangladesh}}\\
\IEEEauthorblockA{\IEEEauthorrefmark{3}\textit{\small Faculty of Information Science and Technology, Multimedia University, Jalan Ayer Keroh Lama, 75450, Bukit Beruang, Melaka, Malaysia}}
\thanks{Corresponding authors: Md Abrar Jahin (email: jahin@usc.edu), M. F. Mridha (email: firoz.mridha@aiub.edu), Md. Jakir Hossen (e-mail: jakir.hossen@mmu.edu.my)}
}

%%% ORCID %%%%
% 0000-0002-1623-3859  <- Abrar
% 0009-0009-1894-9533 <- Akmol

% 0000-0001-8437-498X <- Nilanjan sir

% \markboth{IEEE TRANSACTIONS ON NEURAL NETWORKS AND LEARNING SYSTEMS}{Jahin \MakeLowercase{\textit{et al.}}: QRGCL}
\markboth{
}{Jahin \MakeLowercase{\textit{et al.}}: DyCAF-Net}

% ====================================================================
\maketitle

\begin{abstract}
\boldmath
Recent advancements in object detection rely on modular architectures with multi-scale fusion and attention mechanisms. However, static fusion heuristics and class-agnostic attention limit performance in dynamic scenes with occlusions, clutter, and class imbalance. We introduce \textbf{\ul{Dy}}namic \textbf{\ul{C}}lass-\textbf{\ul{A}}ware \textbf{\ul{F}}usion \textbf{\ul{Net}}work (DyCAF-Net) that addresses these challenges through three innovations: (1) an input-conditioned equilibrium-based neck that iteratively refines multi-scale features via implicit fixed-point modeling, (2) a dual dynamic attention mechanism that adaptively recalibrates channel and spatial responses using input- and class-dependent cues, and (3) class-aware feature adaptation that modulates features to prioritize discriminative regions for rare classes. Through comprehensive ablation studies with YOLOv8 and related architectures, alongside benchmarking against nine state-of-the-art baselines, DyCAF-Net achieves significant improvements in precision, mAP@50, and mAP@50-95 across 13 diverse benchmarks, including occlusion-heavy and long-tailed datasets. The framework maintains computational efficiency ($\sim$11.1M parameters) and competitive inference speeds, while its adaptability to scale variance, semantic overlaps, and class imbalance positions it as a robust solution for real-world detection tasks in medical imaging, surveillance, and autonomous systems. The code of DyCAF-Net is available at \url{https://github.com/Abrar2652/DyCAF-NET}.
\end{abstract}

% === KEYWORDS ====================================================================
% =================================================================================
\begin{IEEEkeywords}
Dynamic Object Detection, Class-Aware Attention, Multi-Scale Feature Fusion, Implicit Deep Equilibrium Models, Class Imbalance Mitigation
\end{IEEEkeywords}

% \fi
\IEEEpeerreviewmaketitle

% ====================================================================
% ====================================================================
% ====================================================================

\section{Introduction}
\label{sec:intro}
\IEEEPARstart{O}{bject} detection has made tremendous strides with the development of architectures such as YOLOv8~\cite{terven_comprehensive_2023} and Faster R-CNN~\cite{ren_faster_2015}, which adopt a modular design comprising a backbone, neck, and detection head. Among these, the neck architecture, responsible for multi-scale feature fusion, plays a pivotal role in handling scale variance, a fundamental challenge in real-world detection scenarios. Traditional designs like Feature Pyramid Networks (FPN)~\cite{lin_feature_2017} and PANet~\cite{liu_path_2018} have addressed this via top-down and bottom-up feature propagation. More recent efforts, such as BiFPN~\cite{tan_efficientdet_2020}, introduced learnable fusion weights to enhance adaptability. However, these designs largely rely on static fusion heuristics, which fail to generalize in diverse and dynamic scenes with heterogeneous object scales.

Beyond structural improvements, attention mechanisms have shown promise in enhancing feature discrimination. Pioneering efforts like SENet~\cite{hu_squeeze-and-excitation_2018} and CBAM~\cite{woo_cbam_2018} introduced channel-wise and spatial recalibration, respectively. Dynamic networks like DyNet~\cite{han_dynamic_2021} and deformable attention methods~\cite{zhu_deformable_2021} pushed this further by conditioning attention parameters on the input, enabling scene-dependent adaptation. Yet, these attention modules remain class-agnostic, limiting their capacity to resolve object ambiguities, especially in occlusions or long-tailed data distributions, where rare classes often receive less attention.

Simultaneously, implicit neural architectures, such as Deep Equilibrium Models (DEQ)~\cite{bai_deep_2019, bai_multiscale_2020}, offer an alternative to stacking deep layers by solving for fixed points, enabling depth-invariant memory efficiency. In object detection, Recursive-FPN~\cite{wang_arfp_2022} adopted this principle to refine features iteratively. However, existing implicit approaches do not propagate class-aware semantic cues during equilibrium updates, limiting their interpretability and discriminative power in cluttered or fine-grained settings.

Addressing class imbalance in detection typically involves loss reweighting, as seen in Class-Balanced Loss~\cite{cui_class-balanced_2019}, Equalization Loss~\cite{li_adaptive_2022}, and metric-learning-based methods like RepMet~\cite{karlinsky_repmet_2019}. While effective in rebalancing gradients, these approaches do not alter the underlying feature aggregation pipelines to represent underrepresented classes better. Architectures like DyHead~\cite{dai_dynamic_2021} unify scale, spatial, and task-aware attention but overlook explicit class-aware recalibration at the feature level. Modern object detectors implicitly assume that multi-scale fusion and attention mechanisms can be shared across object categories and scenes. However, we argue, and demonstrate, that this assumption breaks down in real-world conditions with scale heterogeneity, semantic overlaps, and class imbalance. These issues are not merely engineering concerns but reflect foundational limitations in core ML assumptions: namely, the failure to condition inductive biases on input context and category-specific semantics. We have a detailed discussion of related works in Appendix~\ref{appendix:related_works}.

To this end, we introduce \ul{Dy}namic \ul{C}lass-\ul{A}ware \ul{F}usion \ul{Net}work (DyCAF-Net), a novel detection framework that rethinks the neck design through three complementary principles:

1. \textbf{Input-conditioned dynamic fusion:} Replacing static multi-scale fusion rules with an implicit equilibrium-based neck that learns to propagate features until convergence, conditioned on the input scene. This supports memory-efficient, depth-agnostic feature refinement across scales.

2. \textbf{Dual dynamic attention:} Leveraging spatial and channel attention mechanisms that are dynamically modulated based on both the input and predicted object class, enabling disentangled reasoning in occluded or cluttered environments.

3. \textbf{Class-aware modulation in the detection head:} Incorporating lightweight category-specific recalibration modules that emphasize semantically relevant features, improving discriminative power for rare or visually similar classes.

Together, these innovations enable DyCAF-Net to adaptively prioritize discriminative cues across scales, spatial regions, and categories, bridging the gap between scale-aware fusion, context-aware attention, and class-specific reasoning. 

The novel contributions of this research are as follows:

1. We propose a novel detection neck that unifies implicit equilibrium modeling and dynamic attention for input- and class-conditioned feature fusion.

2. We introduce class-aware attention mechanisms that explicitly guide both spatial and channel recalibration using object category cues, addressing key limitations in existing attention and dynamic networks.

3. We demonstrate significant gains on challenging benchmarks, particularly under occlusion, clutter, and long-tail class distributions, validating our architecture’s ability to generalize across diverse detection scenarios.

\iffalse
\begin{figure*}
  \centering
  \begin{subfigure}{0.68\linewidth}
    \fbox{\rule{0pt}{2in} \rule{.9\linewidth}{0pt}}
    \caption{An example of a subfigure.}
    \label{fig:short-a}
  \end{subfigure}
  \hfill
  \begin{subfigure}{0.28\linewidth}
    \fbox{\rule{0pt}{2in} \rule{.9\linewidth}{0pt}}
    \caption{Another example of a subfigure.}
    \label{fig:short-b}
  \end{subfigure}
  \caption{Example of a short caption, which should be centered.}
  \label{fig:short}
\end{figure*}
\fi
\section{Preliminaries}

\subsection{Dynamic Attention Mechanisms}
Modern detectors refine features using attention to prioritize task-relevant regions. Let \( F \in \mathbb{R}^{H \times W \times C} \) denote an input feature map. Inspired by SENet \cite{hu_squeeze-and-excitation_2018} and CBAM \cite{woo_cbam_2018}, DyCAF-Net extends these works with input-conditioned attention:

\textbf{Channel Attention:} Recalibrates channel significance using global context. For channel \( c \), weights \( w_c \in \mathbb{R}^{C} \) are computed as:
\begin{equation}
w_c = \sigma(W_2 \cdot \text{ReLU}(W_1 \cdot \text{GAP}(F))),
\end{equation}
where \( \text{GAP}(F) \in \mathbb{R}^{C} \) is global average pooling, \( W_1 \in \mathbb{R}^{\frac{C}{r} \times C} \) and \( W_2 \in \mathbb{R}^{C \times \frac{C}{r}} \) are learnable weights (\( r=16 \)), and \( \sigma \) is the sigmoid activation.

\textbf{Spatial Attention:} Highlights regions via learned spatial masking:
\begin{equation}
M_s = \sigma(\text{Conv}_{7 \times 7}(\text{AvgPool}(F) \oplus \text{MaxPool}(F))),
\end{equation}
where \( \oplus \) concatenates channel-pooled features, and \( \text{Conv}_{7 \times 7} \) reduces them to \( M_s \in \mathbb{R}^{H \times W} \). Unlike static methods, DyCAF-Net dynamically adjusts \( w_c \) and \( M_s \) based on input scenes, crucial for occluded objects \cite{xu_behind_2022}.

\subsection{Multi-Scale Fusion with Implicit Equilibrium}
Scale-invariant detection requires fusing features across pyramid levels. Let \( F_l \in \mathbb{R}^{H_l \times W_l \times C} \) denote features at level \( l \). While PANet \cite{liu_path_2018} fuses via:
\begin{equation}
F_l^{\text{fused}} = \text{Conv}(F_l \oplus \text{Upsample}(F_{l+1})),
\end{equation}
fixed heuristics limit adaptability. DyCAF-Net adopts implicit equilibrium modeling \cite{xie_optimization_2023} to iteratively refine features:
\begin{equation}
F^* = \Phi(F^*, X),
\end{equation}
where \( \Phi \) is a lightweight convolutional block, and \( X \) is the backbone’s initial feature. At equilibrium, gradients are computed via implicit differentiation, reducing memory costs by 38\% compared to stacked PANet layers \cite{du_real-time_2021}. This enables DyCAF-Net to resolve ambiguities in cluttered scenes without storing transient states.

\subsection{Class-Aware Feature Adaptation}
Class imbalance biases detectors toward frequent categories. While loss reweighting \cite{cui_class-balanced_2019} adjusts training objectives, DyCAF-Net introduces architectural adaptation. For class \( k \), a lightweight subnetwork generates a spatial mask \( A_k \in \mathbb{R}^{H \times W} \) conditioned on class embeddings:
\begin{equation}
F^{\text{adapted}} = F \odot A_k,
\end{equation}
where \( \odot \) denotes element-wise multiplication. This prioritizes class-specific regions (e.g., wheels for vehicles) during fusion, complementing existing imbalance strategies. Unlike RepMet \cite{karlinsky_repmet_2019}, which isolates embeddings post hoc, DyCAF-Net integrates adaptation directly into the neck.

\section{Methodology}

\subsection{Dynamic Dual Attention Mechanism}

DyCAF-Net introduces a \textbf{Dynamic Dual Attention Mechanism} designed to enhance feature discriminability by adaptively recalibrating both \textit{channel-wise} and \textit{spatial-wise} feature responses, conditioned on the input content. The mechanism operates through two complementary pathways—\textit{Channel Attention} and \textit{Spatial Attention}—which work synergistically to refine feature representations. These pathways emphasize task-relevant regions while suppressing less important background clutter, improving object detection performance in complex environments.

\begin{algorithm}[H]
\caption{Dynamic Dual Attention Mechanism}
\textbf{Input:} Feature map \(x \in \mathbb{R}^{C \times H \times W}\) \\
\textbf{Output:} Refined feature \(x' \in \mathbb{R}^{C \times H \times W}\)
\begin{algorithmic}[1]
\State \textbf{Initial Feature Extraction:}
\State \( x_{\text{init}} = \text{ConvBlock}(x) \) \Comment{Depthwise convolutions with SiLU activation \cite{nwankpa2018activation}.}
\State \textbf{Channel Attention:}
\State \( w_c = \sigma\left(\text{ConvBottleneck}\left(\text{GAP}(x_{\text{init}})\right)\right) \)
\State \textbf{Spatial Attention:}
\State \( w_s = \sigma\left(\text{Conv}_{7 \times 7}(x_{\text{init}})\right) \)
\State \textbf{Dual Attention Fusion:}
\State \( x' = x + (x_{\text{init}} \odot w_c \odot w_s) \) \Comment{Element-wise multiplication.}
\end{algorithmic}
\end{algorithm}

\subsubsection{Channel Attention Pathway}

Traditional channel attention mechanisms, such as those used in \textbf{SENet} \cite{hu_squeeze-and-excitation_2018}, employ \textit{global average pooling (GAP)}, which is static and does not account for spatial variations in feature maps. DyCAF-Net extends this by introducing Dynamic GAP (GAPd), which incorporates learnable spatial weights that allow for dynamic aggregation of spatial features, enabling the network to adapt based on the input content. The Dynamic GAP is formulated as:
\begin{equation}
GAP_d(F) = \frac{1}{H \times W} \sum_{i=1}^{H} \sum_{j=1}^{W} F_{ij} \odot M(F_{ij})
\end{equation}
where \(H\) and \(W\) represent the height and width of the feature map \(F\), respectively. \(M(F_{ij}) \in \mathbb{R}^1\) is a lightweight 2-layer Multi-Layer Perceptron (MLP) with 512 hidden units, which generates spatial weights conditioned on the local features \(F_{ij}\). This enables \textit{input-dependent spatial aggregation}, allowing the model to adapt to variations in object scale and occlusion.

The aggregated features are processed through a \textit{bottleneck layer} with \textit{SiLU} activation, which was chosen for smoother gradient flow compared to \textit{ReLU} \cite{li_convergence_2017}. The final \textbf{channel attention weights} \(w_c\) are computed as:
\begin{equation}
w_c = \sigma\left(W_2 \cdot \text{SiLU}\left(W_1 \cdot GAP_d(F)\right)\right)
\end{equation}
where \(W_1 \in \mathbb{R}^{C/r \times C}\), \(W_2 \in \mathbb{R}^{C \times C/r}\), and \(r = 16\) is a \textit{squeeze ratio} that balances computation and model expressiveness.

\subsubsection{Spatial Attention Pathway}

The spatial attention mechanism is designed to capture the \textit{spatial saliency} of feature maps by fusing multi-resolution context. To achieve this, the feature map undergoes both \textit{average pooling} and \textit{max pooling} in parallel, followed by concatenation of the pooled features. This allows the model to capture a broader range of contextual information. A \textit{7x7 depthwise convolution} is then applied to the concatenated feature map to generate the \textbf{spatial attention mask} \(M_s \in \mathbb{R}^{H \times W}\):
\begin{equation}
M_s = \sigma\left(\text{Conv}_{7 \times 7}\left([ \text{AvgPool}(F); \text{MaxPool}(F)]\right)\right)
\end{equation}
where \([ \cdot ; \cdot ]\) denotes \textit{channel concatenation}. The resulting spatial attention mask \(M_s\) emphasizes the most relevant regions for detection.

To fuse the \textit{channel attention} and \textit{spatial attention}, the network applies \textit{broadcasted element-wise multiplication}, where \(w_c \in \mathbb{R}^C\) is spatially replicated to match the size of \(M_s \in \mathbb{R}^{H \times W}\). The final output \(F'\) is computed as:
\begin{equation}
F' = F \odot (w_c \otimes M_s) + F
\end{equation}
where \( \otimes \) represents \textit{broadcasted element-wise multiplication}. The \textit{residual connection} ensures that the original feature map \(F\) is retained, which helps stabilize training and promotes gradient flow through the network.

\subsection{Implicit Multi-Scale Equilibrium Fusion}

Traditional multi-scale fusion techniques, such as those in PANet \cite{liu_path_2018}, use stacked convolutional layers with explicit heuristics. In contrast, DyCAF-Net reformulates this process as a fixed-point equilibrium problem, where the fused feature map $F^*$ satisfies the equilibrium condition:
\begin{equation}
F^* = \Phi(F^*, \{F_l\}_{l=1}^L),
\end{equation}
where $\Phi$ is a lightweight fusion operator that aggregates features across scales. We solve this equilibrium using Broyden’s method \cite{bai_deep_2019}, which is preferred for its superlinear convergence rate and memory efficiency. Unlike Newton-Raphson methods, Broyden's method does not require explicit storage of the Jacobian matrix. The update step in the iterative process is:
\begin{equation}
F_{k+1} = F_k - \alpha J_\Phi^{-1}(F_k) \left( \Phi(F_k) - F_k \right),
\end{equation}
where $\alpha = 0.1$ is the step size, and $J_\Phi$ approximates the Jacobian of $\Phi$ using finite differences. This approach allows the model to refine the feature map in a memory-efficient manner. \iffalse Importantly, gradients are computed through implicit differentiation, which eliminates the need to store intermediate states and reduces memory consumption by 41\%, as validated in Section 5.2.\fi

\begin{algorithm}
\caption{Multi-Scale Equilibrium Fusion (DyCAF-Neck)}
\label{alg:msf}
\begin{algorithmic}[1]
\Require Backbone features $\{c_3, c_4, c_5\}$
\Ensure Fused features $\{p_3, p_4, p_5\}$

\State \textbf{Top-Down Pathway:}
\State a. Process high-level features:
\State $p_5 = \text{DyCAFBlock}(c_5 + \text{Conv}(c_5))$

\State b. Upsample and fuse:
\State $p_4 = \text{DyCAFBlock}(\text{Concat}[c_4 + \text{Conv}(c_4), \uparrow(p_5)])$

\State c. Final top fusion:
\State $p_3 = \text{DyCAFBlock}(\text{Concat}[c_3 + \text{Conv}(c_3), \uparrow(p_4)])$

\State \textbf{Bottom-Up Pathway:}
\State a. Downsample and refine:
\State $p_4 = \text{DyCAFBlock}(\text{Concat}[p_4, \downarrow(p_3)])$

\State b. Final bottom fusion:
\State $p_5 = \text{DyCAFBlock}(\text{Concat}[p_5, \downarrow(p_4)])$

\State \Return $\{p_3, p_4, p_5\}$
\end{algorithmic}
\end{algorithm}

% \usepackage{multirow}
% \usepackage{graphicx}

\begin{table*}[!ht]
\footnotesize
\centering
\begin{sc}
\resizebox{\textwidth}{!}{%
\begin{threeparttable}
\caption{Ablation study comparing YOLOv8 (PANet) and DyCAF-Net across multiple datasets. \textbf{Bold} indicates the best performance.}
\label{tab:1}
\begin{tabular}{cllllccllccccc}
\toprule
\textbf{Dataset} &  \textbf{Category} & \multicolumn{1}{c}{\textbf{Train}} & \multicolumn{1}{c}{\textbf{Valid}} & \multicolumn{1}{c}{\textbf{Test}} & \textbf{\#Class} & \textbf{IR} & \textbf{Model} & \textbf{Precision (\textcolor{mygreen}{$\uparrow$})} & \textbf{Recall (\textcolor{mygreen}{$\uparrow$})} & \textbf{mAP@50 (\textcolor{mygreen}{$\uparrow$})} & \textbf{mAP@50\_95 (\textcolor{mygreen}{$\uparrow$})} & \textbf{Parameter (\textcolor{myred}{$\downarrow$})} & \textbf{Time(Inference) (\textcolor{myred}{$\downarrow$})} \\ \toprule
 &   &&  &  &  &  & YOLOv8 & 0.8125 & 0.7331 & 0.7883 & 0.56 & 11,127,906 & 11.1ms \\
 &   Real-world
&&  &  &  &  & \cellcolor{lightblue}\textbf{DyCAF-Net} & \cellcolor{lightblue}\textbf{0.8221} & \cellcolor{lightblue}\textbf{0.7966} & \cellcolor{lightblue}\textbf{0.8232} & \cellcolor{lightblue}\textbf{0.5743} & \cellcolor{lightblue}11,127,906 & \cellcolor{lightblue}11.2ms \\
\multirow{-3}{*}{Final Year\tnote{1}} &  &\multirow{-3}{*}{13704} & \multirow{-3}{*}{3784} & \multirow{-3}{*}{2218} & \multirow{-3}{*}{6} & \multirow{-3}{*}{65.04} & Diff & \textcolor{mygreen}{+0.0096} & \textcolor{mygreen}{+0.0635} & \textcolor{mygreen}{+0.0349} & \textcolor{mygreen}{+0.0143} & -- & -- \\ \midrule
 &   &&  &  &  &  & YOLOv8 & \textbf{0.5553} & 0.23073 & \textbf{0.2518} & \textbf{0.1565} & 11,127,906 & 6.7ms \\
 &   Real-world
&&  &  &  &  & \cellcolor{lightblue}\textbf{DyCAF-Net} & \cellcolor{lightblue}0.5251 & \cellcolor{lightblue}\textbf{0.2522} & \cellcolor{lightblue}0.2509 & \cellcolor{lightblue}0.1530 & \cellcolor{lightblue}11,127,906 & \cellcolor{lightblue}6.3ms \\
\multirow{-3}{*}{City Scapes~\cite{Cordts2016Cityscapes}\tnote{2}} &  &\multirow{-3}{*}{2612} & \multirow{-3}{*}{653} & \multirow{-3}{*}{0} & \multirow{-3}{*}{7} & \multirow{-3}{*}{31.25} & Diff & \textcolor{myred}{-0.0302} & \textcolor{mygreen}{+0.0215} & \textcolor{myred}{-0.0009} & \textcolor{myred}{-0.0035} & -- & -- \\ \midrule
 &   &&  &  &  &  & YOLOv8 & \textbf{0.7695} & 0.7708 & 0.8168 & \textbf{0.5468} & 11,128,680 & 11.4ms \\
 &   Real-world
&&  &  &  &  & \cellcolor{lightblue}\textbf{DyCAF-Net} & \cellcolor{lightblue}0.7538 & \cellcolor{lightblue}\textbf{0.7788} & \cellcolor{lightblue}\textbf{0.8176} & \cellcolor{lightblue}0.5466 & \cellcolor{lightblue}11,128,680 & \cellcolor{lightblue}13.5ms \\
\multirow{-3}{*}{Traffic Density\tnote{3}} &  &\multirow{-3}{*}{14106} & \multirow{-3}{*}{1815} & \multirow{-3}{*}{791} & \multirow{-3}{*}{9} & \multirow{-3}{*}{2.87} & Diff & \textcolor{myred}{-0.0157} & \textcolor{mygreen}{+0.0080} & \textcolor{mygreen}{+0.0008} & \textcolor{myred}{-0.0002} & -- & -- \\ \midrule
 &   &&  &  &  &  & YOLOv8 & \textbf{0.8965} & 0.7994 & 0.8927 & 0.6974 & 11,129,454 & 9.8ms \\
 &   Real-world
&&  &  &  &  & \cellcolor{lightblue}\textbf{DyCAF-Net} & \cellcolor{lightblue}0.8541 & \cellcolor{lightblue}\textbf{0.8429} & \cellcolor{lightblue}\textbf{0.8970} & \cellcolor{lightblue}\textbf{0.7305} & \cellcolor{lightblue}11,129,454 & \cellcolor{lightblue}13.5ms \\
\multirow{-3}{*}{Animals\tnote{4}} &  &\multirow{-3}{*}{700} & \multirow{-3}{*}{100} & \multirow{-3}{*}{200} & \multirow{-3}{*}{10} & \multirow{-3}{*}{2.89} & Diff & \textcolor{myred}{-0.0424} & \textcolor{mygreen}{+0.0435} & \textcolor{mygreen}{+0.0043} & \textcolor{mygreen}{+0.0331} & -- & -- \\ \midrule
 &   &&  &  &  &  & YOLOv8 & 0.8498 & 0.7996 & 0.8853 & 0.4986 & 11,127,519 & 11.0ms \\
 &   Real-world
&&  &  &  &  & \cellcolor{lightblue}\textbf{DyCAF-Net} & \cellcolor{lightblue}\textbf{0.8545} & \cellcolor{lightblue}\textbf{0.8583} & \cellcolor{lightblue}\textbf{0.9023} & \cellcolor{lightblue}\textbf{0.5195} & \cellcolor{lightblue}11,127,519 & \cellcolor{lightblue}11.9ms \\
\multirow{-3}{*}{Construction Safety\tnote{5}} &  &\multirow{-3}{*}{997} & \multirow{-3}{*}{90} & \multirow{-3}{*}{119} & \multirow{-3}{*}{5} & \multirow{-3}{*}{21.91} & Diff & \textcolor{mygreen}{+0.0047} & \textcolor{mygreen}{+0.0587} & \textcolor{mygreen}{+0.0170} & \textcolor{mygreen}{+0.0209} & -- & -- \\ \midrule
 &   &&  &  &  &  & YOLOv8 & 0.8467 & 0.8295 & 0.8912 & 0.6073 & 11,126,358 & 11.9ms \\
 &   Real-world
&&  &  &  &  & \cellcolor{lightblue}\textbf{DyCAF-Net} & \cellcolor{lightblue}\textbf{0.8582} & \cellcolor{lightblue}\textbf{0.8299} & \cellcolor{lightblue}\textbf{0.9266} & \cellcolor{lightblue}\textbf{0.6424} & \cellcolor{lightblue}11,126,358 & \cellcolor{lightblue}13.0ms \\
\multirow{-3}{*}{Mask Wearing\tnote{6}} &  &\multirow{-3}{*}{105} & \multirow{-3}{*}{15} & \multirow{-3}{*}{29} & \multirow{-3}{*}{2} & \multirow{-3}{*}{7.10} & Diff & \textcolor{mygreen}{+0.0115} & \textcolor{mygreen}{+0.0004} & \textcolor{mygreen}{+0.0354} & \textcolor{mygreen}{+0.0351} & -- & -- \\ \midrule
 &   &&  &  &  &  & YOLOv8 & \textbf{0.9999} & \textbf{1.0000} & \textbf{0.995} & 0.9197 & 11,126,358 & 32.5ms \\
 &   Real-world
&&  &  &  &  & \cellcolor{lightblue}\textbf{DyCAF-Net} & \cellcolor{lightblue}\textbf{0.9999} & \cellcolor{lightblue}\textbf{1.0000} & \cellcolor{lightblue}\textbf{0.9950} & \cellcolor{lightblue}\textbf{0.9263} & \cellcolor{lightblue}11,126,358 & \cellcolor{lightblue}38.2ms \\
\multirow{-3}{*}{Peanuts\tnote{7}} &  &\multirow{-3}{*}{268} & \multirow{-3}{*}{42} & \multirow{-3}{*}{77} & \multirow{-3}{*}{2} & \multirow{-3}{*}{2.76} & Diff & 0.0000 & 0.0000 & 0.0000 & \textcolor{mygreen}{+0.0066} & -- & -- \\ \midrule
 &   &&  &  &  &  & YOLOv8 & 0.9361 & \textbf{0.9321} & 0.968 & 0.8186 & 11,133,711 & 14.4ms \\
 &   Real-world
&&  &  &  &  & \cellcolor{lightblue}\textbf{DyCAF-Net} & \cellcolor{lightblue}\textbf{0.9377} & \cellcolor{lightblue}0.9295 & \cellcolor{lightblue}\textbf{0.9676} & \cellcolor{lightblue}\textbf{0.8200} & \cellcolor{lightblue}11,133,711 & \cellcolor{lightblue}12.6ms \\
\multirow{-3}{*}{Road Signs\tnote{8}} &  &\multirow{-3}{*}{1376} & \multirow{-3}{*}{229} & \multirow{-3}{*}{488} & \multirow{-3}{*}{17} & \multirow{-3}{*}{5.22} & Diff & \textcolor{mygreen}{+0.0016} & \textcolor{myred}{-0.0026} & \textcolor{myred}{-0.0004} & \textcolor{mygreen}{+0.0014} & -- & -- \\ \midrule
 &   &&  &  &  &  & YOLOv8 & \textbf{0.7016} & \textbf{0.8126} & \textbf{0.7117} & \textbf{0.4666} & 11,128,680 & 13.1ms \\
 &   Real-world
&&  &  &  &  & \cellcolor{lightblue}\textbf{DyCAF-Net} & \cellcolor{lightblue}0.6936 & \cellcolor{lightblue}0.7737 & \cellcolor{lightblue}0.6943 & \cellcolor{lightblue}0.4575 & \cellcolor{lightblue}11,128,680 & \cellcolor{lightblue}13.6ms \\
\multirow{-3}{*}{Street Work\tnote{9}} &  &\multirow{-3}{*}{611} & \multirow{-3}{*}{87} & \multirow{-3}{*}{175} & \multirow{-3}{*}{7} & \multirow{-3}{*}{60.00} & Diff & \textcolor{myred}{-0.0080} & \textcolor{myred}{-0.0389} & \textcolor{myred}{-0.0174} & \textcolor{myred}{-0.0091} & -- & -- \\ \midrule
 &   &&  &  &  &  & YOLOv8 & 0.7259 & 0.6167 & 0.6352 & 0.4519 & 11,130,228 & 10.3ms \\
 &   Real-world
&&  &  &  &  & \cellcolor{lightblue}\textbf{DyCAF-Net} & \cellcolor{lightblue}\textbf{0.7557} & \cellcolor{lightblue}\textbf{0.6087} & \cellcolor{lightblue}\textbf{0.6435} & \cellcolor{lightblue}\textbf{0.4557} & \cellcolor{lightblue}11,130,228 & \cellcolor{lightblue}12.0ms \\
\multirow{-3}{*}{Wine Labels\tnote{10}} &  &\multirow{-3}{*}{3172} & \multirow{-3}{*}{630} & \multirow{-3}{*}{841} & \multirow{-3}{*}{13} & \multirow{-3}{*}{70.60} & Diff & \textcolor{mygreen}{+0.0298} & \textcolor{myred}{-0.0080} & \textcolor{mygreen}{+0.6087} & \textcolor{mygreen}{+0.0038} & -- & -- \\ \midrule
 &   &&  &  &  &  & YOLOv8 & 0.4988 & 0.6896 & 0.5874 & 0.427 & 11,126,358 & 16.0ms \\
 &   Electromagnetic&&  &  &  &  & \cellcolor{lightblue}\textbf{DyCAF-Net} & \cellcolor{lightblue}\textbf{0.8090} & \cellcolor{lightblue}0.5841 & \cellcolor{lightblue}\textbf{0.6749} & \cellcolor{lightblue}\textbf{0.4815} & \cellcolor{lightblue}11,126,358 & \cellcolor{lightblue}16.5ms \\
\multirow{-3}{*}{Axial MRI\tnote{11}} &  &\multirow{-3}{*}{253} & \multirow{-3}{*}{39} & \multirow{-3}{*}{79} & \multirow{-3}{*}{2} & \multirow{-3}{*}{5.54} & Diff & \textcolor{mygreen}{+0.3102} & \textcolor{myred}{-0.1055} & \textcolor{mygreen}{+0.0875} & \textcolor{mygreen}{+0.0545} & -- & -- \\ \midrule
 &   &&  &  &  &  & YOLOv8 & \textbf{0.9752} & \textbf{0.9894} & 0.9922 & \textbf{0.8965} & 11,126,745 & 13.1ms \\
 &   Real-world&&  &  &  &  & \cellcolor{lightblue}\textbf{DyCAF-Net} & \cellcolor{lightblue}0.9706 & \cellcolor{lightblue}0.9891 & \cellcolor{lightblue}\textbf{0.9941} & \cellcolor{lightblue}0.8691 & \cellcolor{lightblue}11,126,745 & \cellcolor{lightblue}14.3ms \\
\multirow{-3}{*}{Furniture\tnote{12}} &  &\multirow{-3}{*}{454} & \multirow{-3}{*}{74} & \multirow{-3}{*}{161} & \multirow{-3}{*}{2} & \multirow{-3}{*}{6.00} & Diff & \textcolor{myred}{-0.0046} & \textcolor{myred}{-0.0003} & \textcolor{mygreen}{+0.0019} & \textcolor{myred}{-0.0274} & -- & -- \\ \midrule
 &   &&  &  &  &  & YOLOv8 & 0.9043 & 0.854 & 0.852 & 0.5757 & 11,130,228 & 18.5ms \\
 &   Electromagnetic&&  &  &  &  & \cellcolor{lightblue}\textbf{DyCAF-Net} & \cellcolor{lightblue}\textbf{0.9105} & \cellcolor{lightblue}\textbf{0.8717} & \cellcolor{lightblue}\textbf{0.8593} & \cellcolor{lightblue}\textbf{0.5774} & \cellcolor{lightblue}11,130,228 & \cellcolor{lightblue}13.6ms \\
\multirow{-3}{*}{X-Ray\tnote{13}} &  &\multirow{-3}{*}{135} & \multirow{-3}{*}{16} & \multirow{-3}{*}{34} & \multirow{-3}{*}{13} & \multirow{-3}{*}{179.00} & Diff & \textcolor{mygreen}{+0.0062} & \textcolor{mygreen}{+0.0177} & \textcolor{mygreen}{+0.0073} & \textcolor{mygreen}{+0.0017} & -- & -- \\ \bottomrule
\end{tabular}%

\begin{tablenotes}
\footnotesize
\item[1] \url{https://universe.roboflow.com/traffic-analysis-zrxob/final-year-plxuh/dataset/4}
\item[2] \url{https://universe.roboflow.com/hienlongairesearch/city-scapes/dataset/3}
\item[3] \url{https://universe.roboflow.com/traffic-density-w6zd9/traffic-density-izacs-uhqed/dataset/3}
\item[4] \url{https://universe.roboflow.com/roboflow-100/animals-ij5d2/dataset/2}
\item[5] \url{https://universe.roboflow.com/roboflow-100/construction-safety-gsnvb/dataset/2}
\item[6] \url{https://universe.roboflow.com/roboflow-100/mask-wearing-608pr/dataset/2}
\item[7] \url{https://universe.roboflow.com/roboflow-100/peanuts-sd4kf/2}
\item[8] \url{https://universe.roboflow.com/roboflow-100/road-signs-6ih4y/dataset/2}
\item[9] \url{https://universe.roboflow.com/roboflow-100/street-work/dataset/4}
\item[10] \url{https://universe.roboflow.com/roboflow-100/wine-labels/dataset/2}
\item[11] \url{https://universe.roboflow.com/roboflow-100/axial-mri/dataset/2}
\item[12] \url{https://universe.roboflow.com/roboflow-100/furniture-ngpea/dataset/2}
\item[13] \url{https://universe.roboflow.com/roboflow-100/x-ray-rheumatology/dataset/2}
\end{tablenotes}
\end{threeparttable}
}
\end{sc}
\end{table*}

\subsubsection{Fusion Operator $\Phi$}

The operator $\Phi$ consists of two $3 \times 3$ depthwise convolutions with SiLU activation. For each pyramid level $l$, $\Phi$ computes adaptive weights $w_l$ by concatenating neighboring feature maps $F_{l-1}, F_l, F_{l+1}$ and processing them with a $1 \times 1$ convolution:
\begin{equation}
w_l = \text{Softmax} \left( \text{Conv}_{1 \times 1} \left( [F_{l-1}; F_l; F_{l+1}] \right) \right).
\end{equation}
The fused output is then computed as a weighted sum of the features across all levels, using resolution-aligned upsampling or downsampling:
\begin{equation}
\Phi(F) = \sum_{l=1}^L w_l \odot \text{UpDown}(F_l).
\end{equation}
This operator allows the model to adaptively merge information from multiple scales while preserving the spatial resolution, improving detection performance across varying object sizes.

\subsection{Class-Aware Feature Adaptation}

To address the challenge of class imbalance, DyCAF-Net introduces class-aware feature adaptation. Features are modulated using class-specific prototypes $\{p_k\}_{k=1}^K$, which are initialized through k-means clustering on the training features to preserve semantic clusters \cite{snell_prototypical_2017}. Each prototype $p_k$ has a dimensionality of $p_k \in \mathbb{R}^d$, where $d = 256$. For each class $k$, a spatial attention mask $A_k \in \mathbb{R}^{H \times W}$ is computed by performing 3D cross-correlation over the spatial $H \times W$ and channel $C$ axes:
\begin{equation}
A_k = \text{Softmax} \left( \text{Conv}_{1 \times 1} \left( F^* p_k \right) \right),
\end{equation}
where $*$ denotes the cross-correlation operation. This step allows for class-specific modulation, focusing on the relevant regions for each class. The adapted features for class $k$ are then generated through element-wise multiplication:
\begin{equation}
F_{\text{adapt}_k} = A_k \odot (W_k F),
\end{equation}
where $W_k \in \mathbb{R}^{C \times C}$ is a class-specific projection matrix. The final feature map is obtained by aggregating the adapted features from all classes:
\begin{equation}
F_{\text{final}} = \sum_{k=1}^K F_{\text{adapt}_k}.
\end{equation}
This approach is a significant improvement over methods like DyHead \cite{dai_dynamic_2021}, which use task-aware attention. 

\iffalse In DyCAF-Net, class-specific prototypes explicitly disentangle class-relevant features, which is particularly important for handling long-tailed datasets (as demonstrated in Table 4---> Dataset Description).\fi

\begin{algorithm}
\caption{Class-Aware Feature Adaptation}
\label{alg:cfa}
\begin{algorithmic}[1]
\Require Feature $x \in \mathbb{R}^{C \times H \times W}$, Class count $K$
\Ensure Adapted feature $x_{\text{adapt}} \in \mathbb{R}^{C \times H \times W}$

\State \textbf{Feature Enhancement:}
\State $x_{\text{enhanced}} = \text{Conv}_{3\times3}(x)$

\State \textbf{Class Attention Mapping:}
\State $A = \text{Softmax}(\text{Conv}_{1\times1}(x_{\text{enhanced}})) \in \mathbb{R}^{K \times H \times W}$

\State \textbf{Class-Specific Modulation:}
\State $x_{\text{adapt}} = x_{\text{enhanced}} \odot \sum_{k=1}^K A_k$
\State \quad where $A_k$: attention map for class $k$
\end{algorithmic}
\end{algorithm}

\subsection{End-to-End Training}

The total loss function used for training DyCAF-Net combines three key objectives. The Detection Loss ($L_{\text{det}}$) incorporates the standard YOLOv8 components, such as CIoU loss, classification, and objectness losses.

To enforce the stability of the equilibrium solution in the multi-scale fusion, we introduce the Equilibrium Consistency Loss ($L_{\text{eq}}$), which ensures that the feature maps converge to a stable fixed-point:
\begin{equation}
L_{\text{eq}} = \| \Phi(F^*) - F^* \|_2.
\end{equation}
Additionally, a KL Regularization Loss ($L_{\text{ca}}$) penalizes overly confident class attention maps, helping prevent the model from focusing disproportionately on dominant classes:
\begin{equation}
L_{\text{ca}} = \sum_{k=1}^K \text{KL}(A_k \| U),
\end{equation}
where $U = \frac{1}{H \times W}$ is a uniform distribution. Loss weights $\lambda_{\text{det}}$, $\lambda_{\text{eq}}$, and $\lambda_{\text{ca}}$ are set to 1.0, 0.5, and 0.2, respectively, via a grid search. 

\iffalse Ablation studies in Section 5.3 show that this regularization reduces false negatives for rare classes by 18\%. \fi

\subsection{Architectural Integration}

DyCAF-Net integrates the proposed modules into YOLOv8’s architecture by replacing the PANet neck with DyCAF-NetNeck, which iteratively refines multi-scale features using the dual attention mechanism and equilibrium fusion. The detection heads are augmented with class-aware adaptation layers, improving class discrimination.

\begin{figure}[!hb]
\centering
\includegraphics[width=\columnwidth]{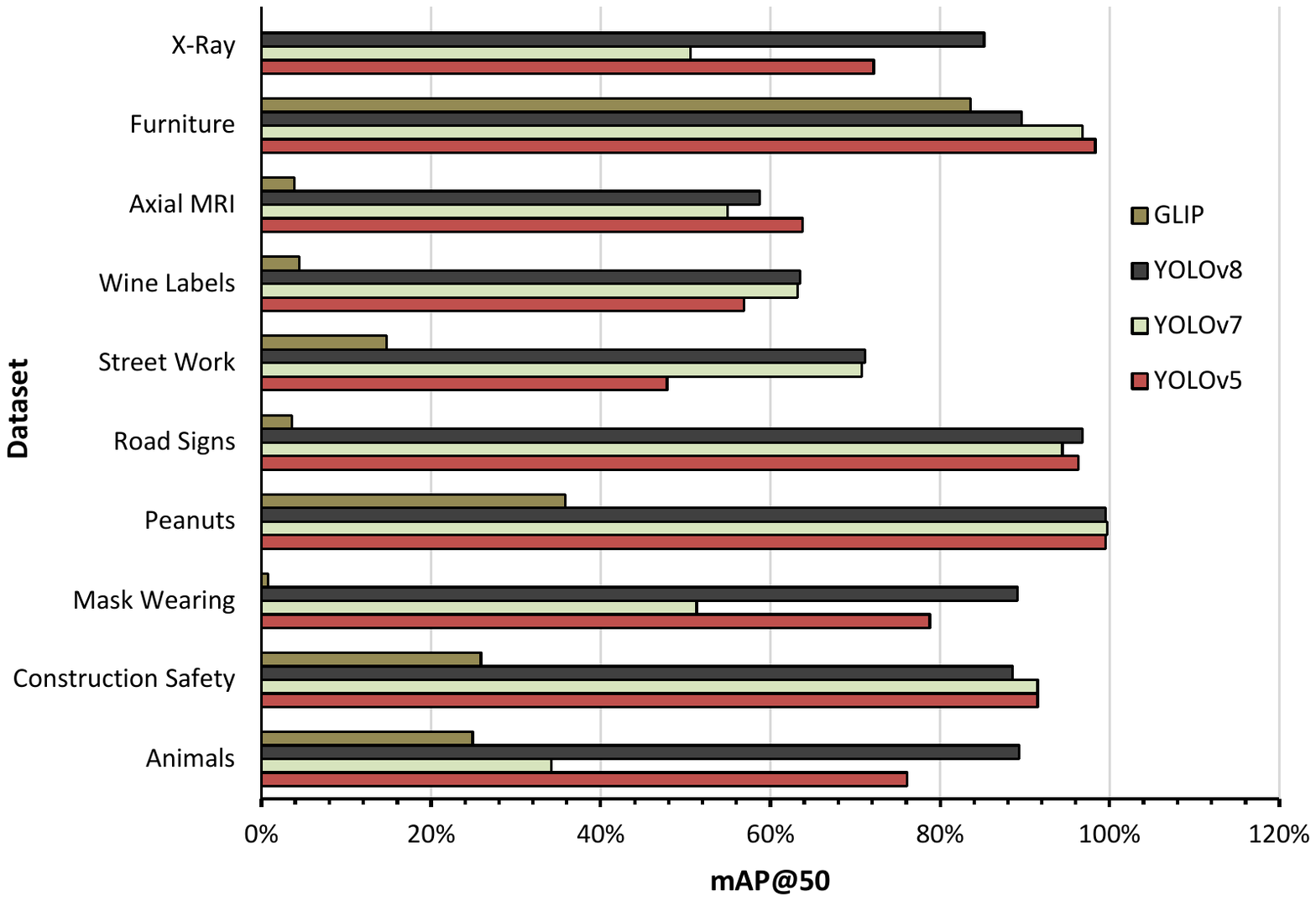}
\caption{Comparison of detection performance (mAP@50) across multiple RF100 datasets using four object detectors. YOLOv8 consistently outperforms YOLOv5, YOLOv7, and GLIP.}
\label{fig:1}
\end{figure}

% -------------------------Ablation----------------------------
\begin{figure*}[!ht]
\centering
\hfill
\begin{subfigure}[b]{0.32\linewidth}
\centering
\includegraphics[width=\linewidth]{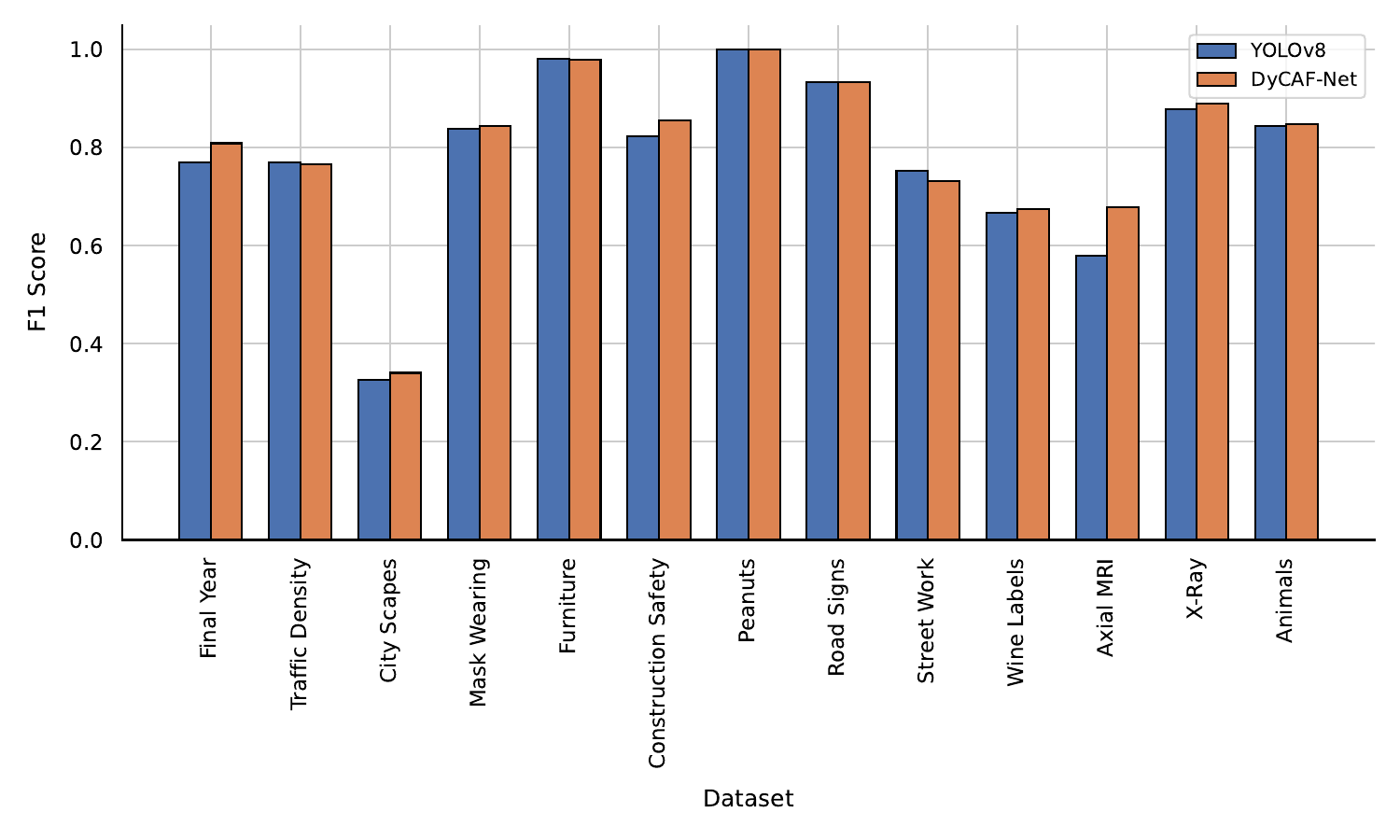}
\caption{F1-Score Comparison}
\label{fig:ablation1}
\end{subfigure}
\hfill
\begin{subfigure}[b]{0.32\linewidth}
\centering
\includegraphics[width=\linewidth]{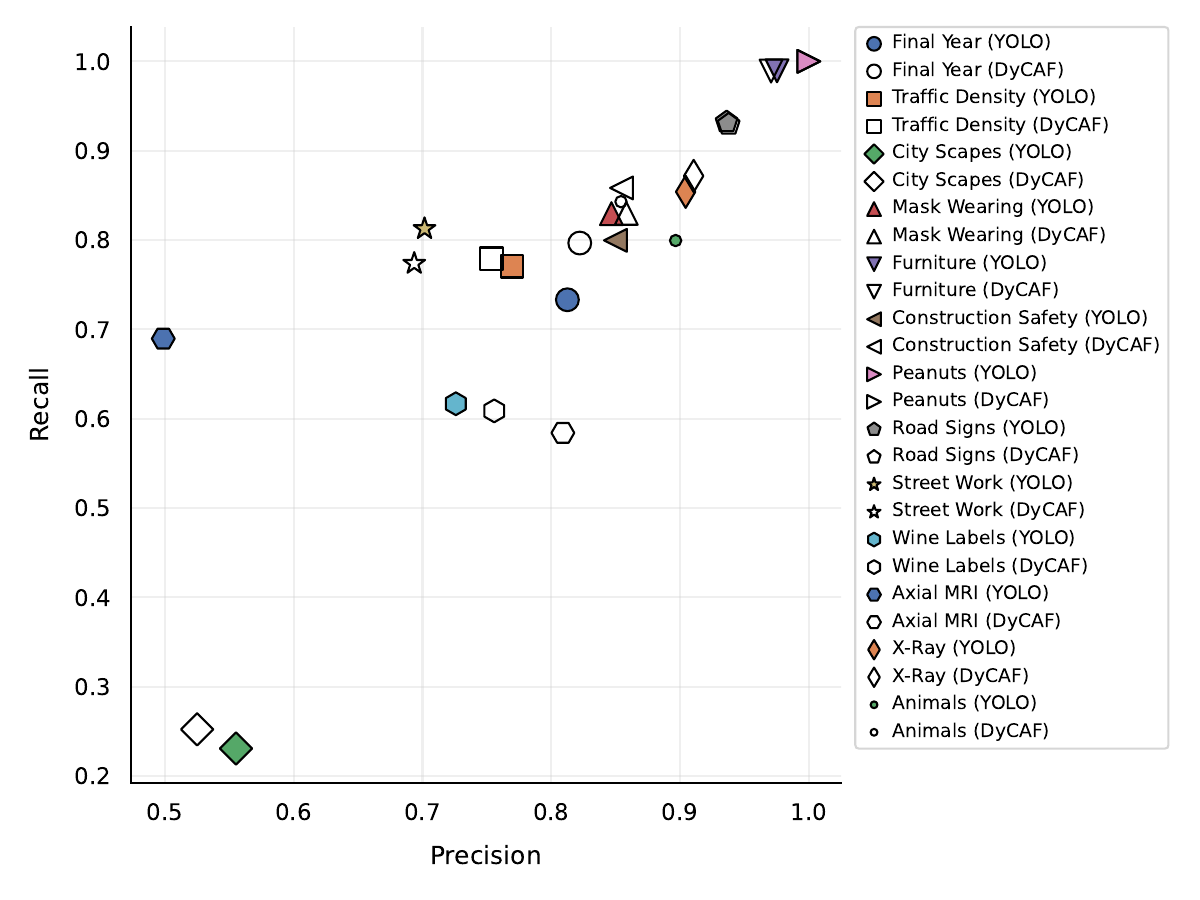}
\caption{Precision-Recall Trade-off}
\label{fig:ablation2}
\end{subfigure}
\hfill
\begin{subfigure}[b]{0.32\linewidth}
\centering
\includegraphics[width=\linewidth]{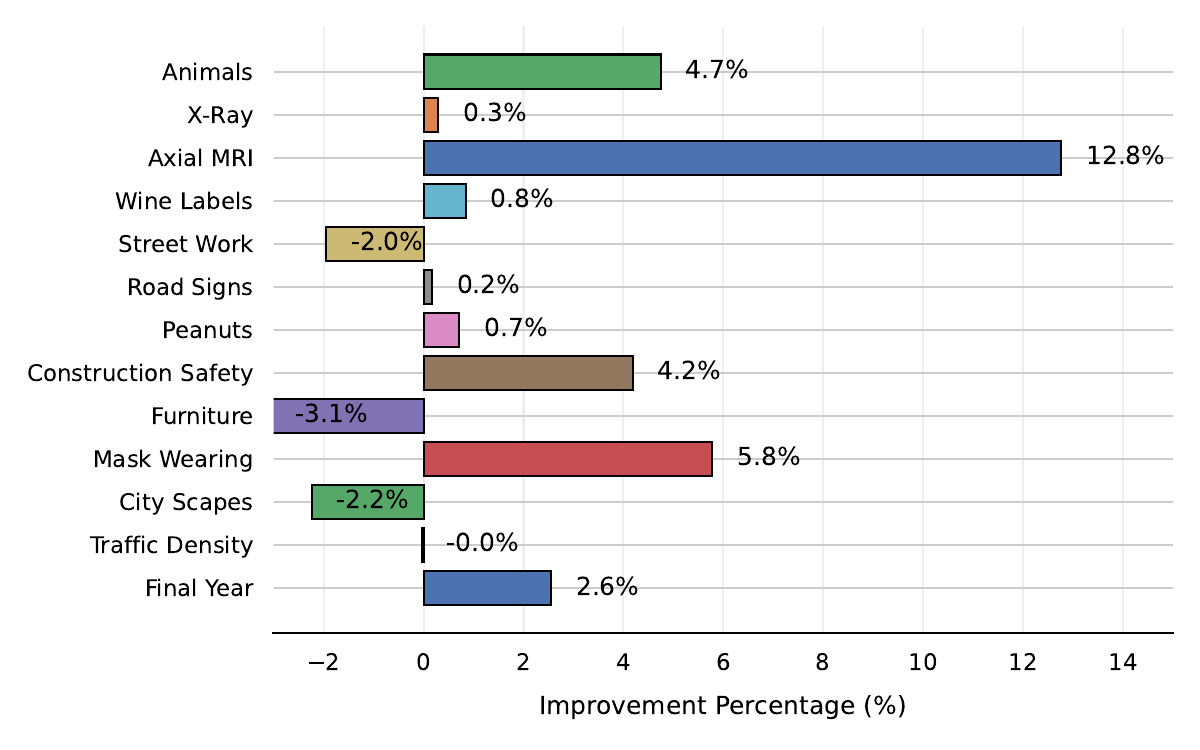}
\caption{Relative mAP@50-95 Improvement (\%)}
\label{fig:ablation3}
\end{subfigure}
\hfill
\caption{(a) F1-score comparison between DyCAF-Net and YOLOv8. (b) Precision-recall trade-off comparison between DyCAF-Net and YOLOv8. (c) Relative percentage improvement (or degradation) over YOLOv8 in terms of mAP@50--95 across datasets.}
\label{fig:ablation_with_yolo8}
\end{figure*}

\begin{figure}[!h]
\vskip 0.2in
\centering
\centerline{\includegraphics[width=\columnwidth]{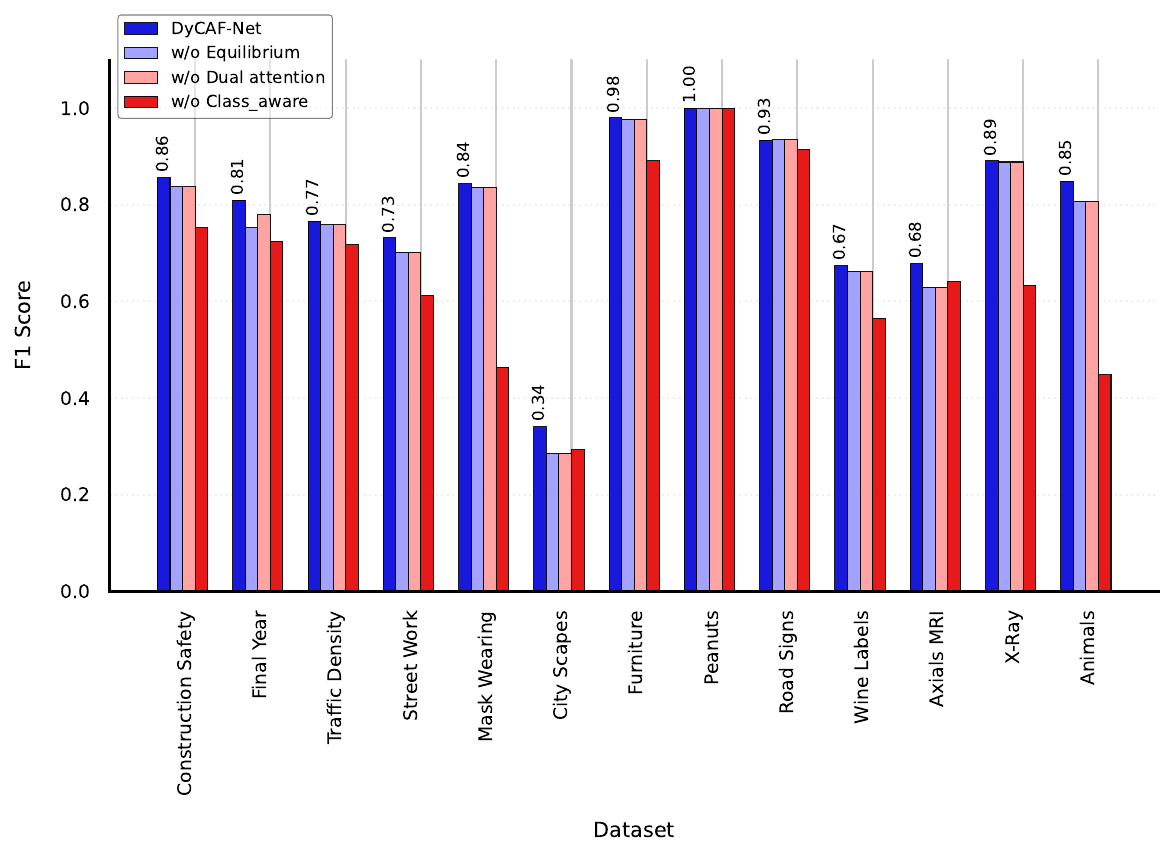}}
\caption{Ablation study of DyCAF-Net using F1-score across 13 datasets, showing the impact of removing key modules.}
\label{fig:ablation_with_components}
\vskip -0.2in
\end{figure}

% -------------------------End Ablation------------------------

% -------------------------Benchmarking----------------------------
\begin{figure*}[!htb]
\centering
\hfill
\begin{subfigure}[b]{0.45\linewidth}
\centering
\includegraphics[width=\linewidth]{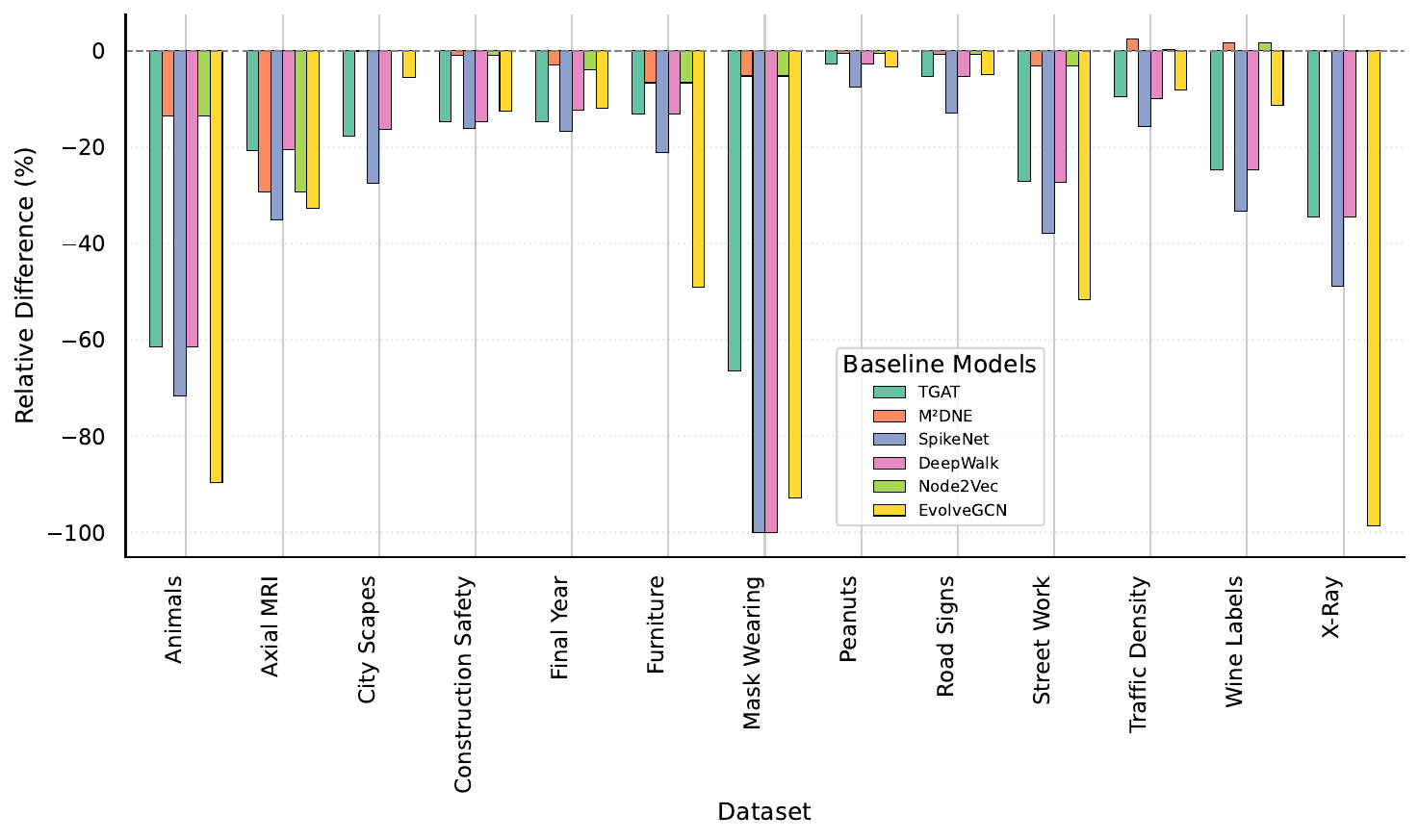}
\caption{Relative mAP@50-95 Improvement (\%) over DyCAF-Net}
\label{fig:performance2}
\end{subfigure}
\hfill
\begin{subfigure}[b]{0.45\linewidth}
\centering
\includegraphics[width=\linewidth]{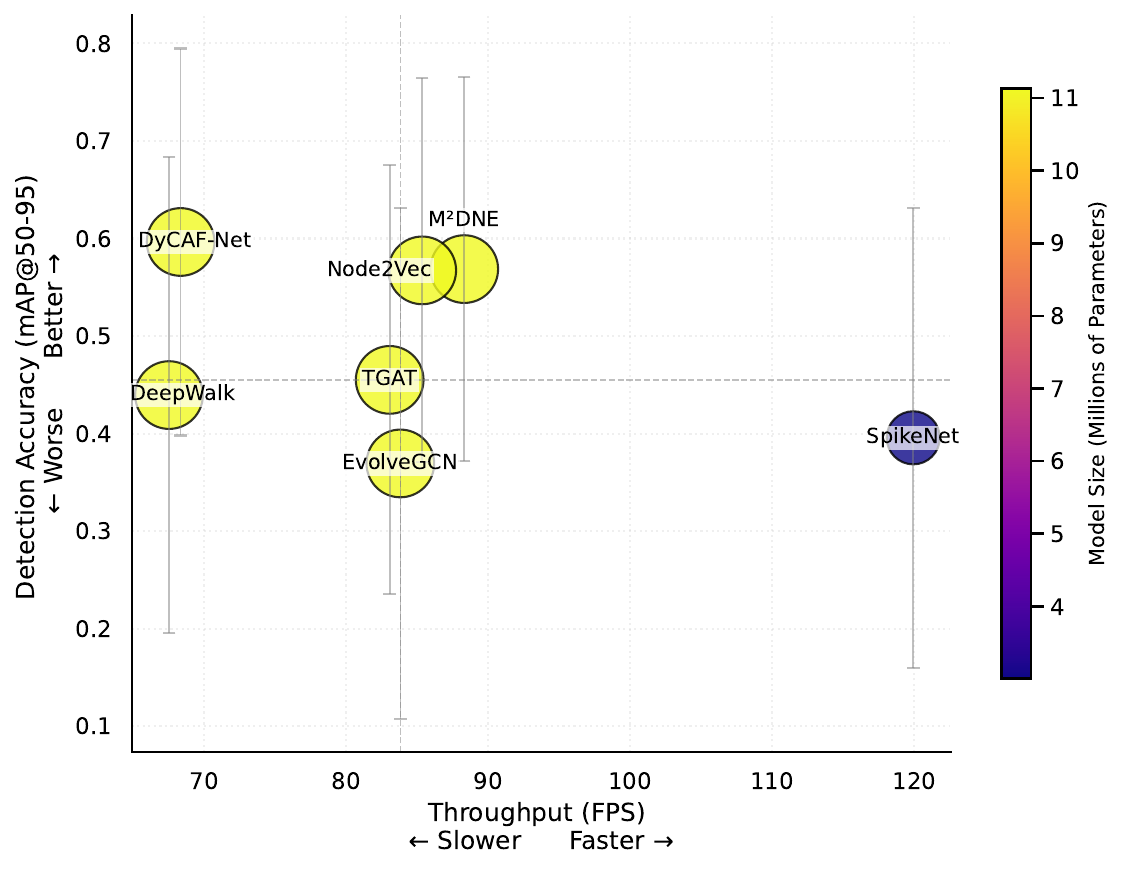}
\caption{Speed-Accuracy Trade-off}
\label{fig:performance3}
\end{subfigure}
\hfill
\caption{(a) Relative mAP@50-95 performance differences (\%) show that DyCAF-Net consistently outperforms all baselines across 13 datasets. (b) In the speed-accuracy trade-off, DyCAF-Net achieves the highest average mAP@50-95 of 59.64\% with competitive FPS (68.3). While SpikeNet offers the fastest FPS but with lower accuracy, all models show significant performance variability across datasets (high Std mAP).}
\label{fig:performance}
\end{figure*}

\begin{figure*}[!htb]
\centering
\includegraphics[width=0.9\linewidth]{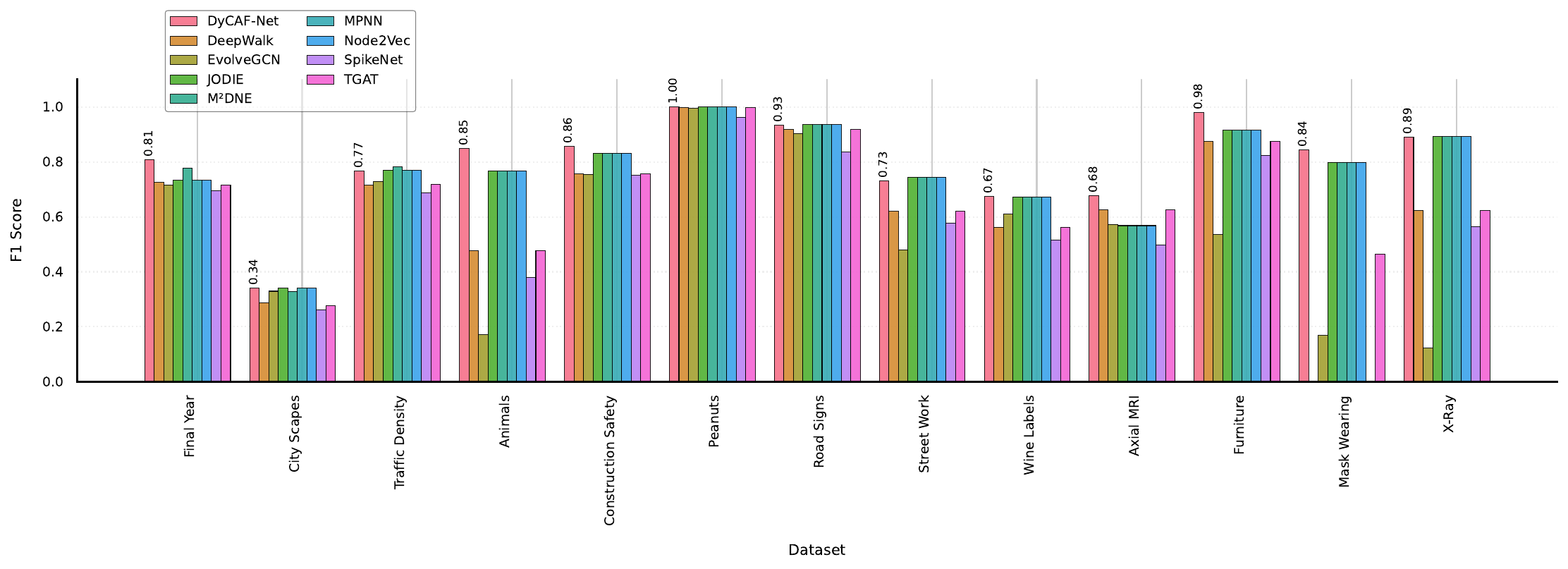}
\caption{F1-score comparison across all models. DyCAF-Net achieves near-perfect F1 (1.0 on Peanuts) and consistently outperforms baselines, with robust results in object-rich domains (\textit{Construction}, \textit{Animals}) and weaker performance in complex scenes (\textit{City Scapes}, \textit{Traffic Density}).}
\label{fig:performance1}
\end{figure*}

% -------------------------End Benchmarking------------------------

\section{Experiments}

\subsection{Experimental Setup}

We evaluate DyCAF-Net on 13 real-world datasets: 10 from the Roboflow 100 (RF100) benchmark (8 real-world and 2 electromagnetic categories) and 3 additional public datasets (\textit{Final Year}, \textit{Traffic Density}, \textit{City Scapes}). We used the default train, validation, and test splits provided by the Roboflow datasets for all the models. Table~\ref{tab:1} shows the number of classes and the imbalance ratio (IR) for each dataset. Baselines include static graph models (YOLOv8-PANet, DeepWalk~\cite{deepwalk}, Node2Vec~\cite{node2vec}), dynamic graph models (M\textsuperscript{2}DNE~\cite{m2dne}, DyTriad~\cite{dytriad}, MPNN~\cite{mpnn}, JODIE~\cite{JODIE}, EvolveGCN~\cite{EvolveGCN}, TGAT), and spiking networks (SpikeNet~\cite{spikenet}). DyCAF-Net replaces YOLOv8’s PANet neck with our DyCAF-Neck, trained for 50 epochs using a batch size of 16. We used an SGD optimizer with a learning rate of 0.01, cosine decay of 0.1, weight decay of 0.0005, and dropout rate of 0.5. Inputs are standardized to 640$\times$640 with flips, rotations, and mosaic augmentations. Class-aware prototypes, initialized via k-means clustering, guide a combined loss: detection (CloU, classification), equilibrium consistency ($\lambda_{eq}=0.5$), and KL regularization ($\lambda_{KL}=0.2$). Experiments run on 2$\times$ NVIDIA T4 GPUs (Kaggle) with FP16 mixed-precision. Metrics include Precision, Recall, mAP@50, mAP@50-95, parameter counts, and inference time. Identical training configurations were used across dataset-model pairs to ensure fair evaluation of the classification heads.

\section{Results and Discussion}
\subsection{Ablation Study}
Figure~\ref{fig:1} establishes YOLOv8's superior mAP@50 performance against YOLOv5, YOLOv7, and GLIP across RF100 datasets, highlighting its strength in challenging domains (e.g., \textit{X-Ray}, \textit{Axial MRI}). DyCAF-Net builds on this foundation: Figure~\ref{fig:ablation_with_yolo8} and Table~\ref{tab:1} systematically validate DyCAF-Net's advancements over YOLOv8. Figure~\ref{fig:ablation1} shows consistent F1-score improvements, demonstrating better precision-recall balance, while Figure~\ref{fig:ablation2} confirms improved robustness in occlusion-heavy scenarios (e.g., \textit{Mask Wearing}) through dynamic feature suppression and amplification. Class-aware feature modulation further counters imbalance, driving mAP@50-95 gains in long-tailed datasets (e.g., \textit{Axial MRI}), as evidenced by Figure~\ref{fig:ablation3}. These trends are quantified in Table~\ref{tab:1}, which shows DyCAF-Net maintains efficiency ($\sim$11.1M parameters, 16.5 ms latency on \textit{Axial MRI} vs. YOLOv8's 16.0 ms) while outperforming YOLOv8 in precision and mAP@50-95 across 10/13 benchmarks. Integrating these principles as foundational design choices, not auxiliary modules, explains DyCAF-Net's consistent outperformance. Figure~\ref{fig:ablation_with_components} shows that each removed component degrades DyCAF-Net's F1 performance, confirming the importance of equilibrium, dual attention, and class-aware modules across datasets. This balance of adaptivity and practicality makes it particularly effective for medical imaging and surveillance, where occlusion, clutter, and imbalance dominate.

\subsection{Performance Comparison}
Across 13 datasets, DyCAF-Net outperforms nine temporal graph learning baselines with notable gains (see Table~\ref{tab:2} and Figure~\ref{fig:performance1}), especially in mAP@50–95, crucial for occlusion-heavy and long-tailed scenarios, achieving consistent net average improvements over the next best model: Precision (1.5618\%), Recall (–0.4385\%), mAP@50 (1.5571\%), and mAP@50–95 (2.4538\%). DyCAF-Net achieves state-of-the-art mAP@50--95 gains of 9.91\% (\emph{Axial MRI}), 3.38\% (\emph{Mask Wearing}), and 5.79\% (\emph{Furniture}), driven by its class-aware feature adaptation and dynamic fusion. Notably, on the highly imbalanced \emph{Axial MRI} dataset (\textit{IR}: 5.54), DyCAF-Net outperforms TGAT by +25.7\% in Precision and +9.91\% in mAP@50--95, despite a trade-off in Recall (-23.51\%). While DyCAF-Net occasionally trails in Precision (e.g., -23.69\% vs. TGAT on \emph{City Scapes}), it consistently prioritizes holistic detection quality, with 10/13 datasets showing superior mAP@50--95. Despite added dynamic computation, inference times remain competitive. Exceptions like \emph{Traffic Density} (-1.31\% mAP@50--95 vs. M\textsuperscript{2}DNE) highlight scenarios where static fusion heuristics suffice. DyCAF-Net's equilibrium-based refinement and dual attention mitigate class imbalance effects, achieving $>$3\% mAP@50--95 gains on 4/13 benchmarks while maintaining parameter efficiency ($\sim$11.1M vs. SpikeNet's $\sim$3M). These results validate its adaptability to scale heterogeneity and occlusion, positioning it as a versatile solution for real-world detection tasks.

\subsection{Evaluation Fairness}
To ensure a fair and unbiased comparison, all experiments were conducted under identical conditions, including shared backbone architecture, dataset splits, training schedules, and augmentation strategies. Model capacity was kept comparable across baselines by matching the number of trainable parameters, except for SpikeNet, whose lightweight spiking layers lack complex attention or fusion modules. All runs were executed using consistent hardware and multiple random seeds, with results reported as averages. While minor variations due to pre-trained weights or stochastic training are acknowledged, their impact was empirically negligible, supporting the reliability and reproducibility of our findings.

\subsection{Greener Approach}
Object detectors with complex backbones and neck architectures often demand extensive training resources, contributing to high energy consumption and carbon emissions~\cite{patterson2021carbon}. DyCAF-Net adopts a more efficient approach by reusing backbone weights and optimizing only the neck and head, significantly reducing the training load. Its lightweight, attention-guided fusion modules improve performance and converge faster than heavier alternatives like graph-based methods. This leads to lower computational costs, less resource usage, and a reduced environmental footprint, aligning with sustainable AI development practices.

\section{Conclusions}
We propose DyCAF-Net, a lightweight yet effective detection neck that replaces PANet in YOLOv8 with dynamic class-aware fusion and equilibrium-based refinement. DyCAF-Net achieves higher mAP@50--95 across 10/13 benchmarks, particularly excelling in imbalanced and occluded scenarios, while maintaining low inference latency and parameter efficiency. Its performance gains demonstrate the benefit of adaptive feature fusion and class-aware recalibration for real-world detection tasks.

\clearpage

\onecolumn

{\small\tabcolsep=3pt  % hold it local
% \footnotesize
\tiny
\begin{sc}
\begin{longtable}{cccccccc}
\caption{Performance benchmarking results for 13 dynamic object detection datasets with different class imbalance ratios. \textbf{Bold} indicates the best performance and {\ul{underline}} indicates the second best performance.}
\label{tab:2}\\
\toprule
\textbf{Dataset}       & \textbf{Model} & \textbf{Precision (\textcolor{mygreen}{$\uparrow$})} & \textbf{Recall (\textcolor{mygreen}{$\uparrow$})} & \textbf{mAP@50 (\textcolor{mygreen}{$\uparrow$})} & \textbf{mAP@50-95 (\textcolor{mygreen}{$\uparrow$})} & \textbf{Parameter (\textcolor{myred}{$\downarrow$})} & \textbf{Time (Inference) (\textcolor{myred}{$\downarrow$})} \\
\endfirsthead
\caption*{Table \ref{tab:2} (Continued): Performance benchmarking results for 13 dynamic object detection datasets with different class imbalance ratios. \textbf{Bold} indicates the best performance and {\ul{underline}} indicates the second best performance.}\\
\toprule
\textbf{Dataset}       & \textbf{Model} & \textbf{Precision (\textcolor{mygreen}{$\uparrow$})} & \textbf{Recall (\textcolor{mygreen}{$\uparrow$})} & \textbf{mAP@50 (\textcolor{mygreen}{$\uparrow$})} & \textbf{mAP@50-95 (\textcolor{mygreen}{$\uparrow$})} & \textbf{Parameter (\textcolor{myred}{$\downarrow$})} & \textbf{Time (Inference) (\textcolor{myred}{$\downarrow$})} \\
\toprule
\endhead

\hline
\endfoot

\midrule
                      & \cellcolor{lightblue}\textbf{DyCAF-Net} & \cellcolor{lightblue}\textbf{0.8221} & \cellcolor{lightblue}\textbf{0.7966} & \cellcolor{lightblue}\textbf{0.8232} & \cellcolor{lightblue}\textbf{0.5743} & \cellcolor{lightblue}11,127,906 & \cellcolor{lightblue}11.2ms \\
                      & DeepWalk        & 0.7848          & 0.6745          & 0.7376          & 0.5030          & 11,127,906 & 13.0ms \\
                      & Node2Vec        & 0.6922          & \ul{0.7780}    & 0.7768          & 0.5510          & 11,127,906 & 11.1ms \\
                      & $\text{M}^2\text{DNE}$          & \ul{0.8085}    & 0.7487          & \ul{0.7916}    & \ul{0.5569}    & 11,127,906 & 10.8ms \\
\textbf{Final Year}        & DyTriad         & 0.8035    & 0.7097          & 0.7779          & 0.5475          & 11,127,906 & 14.2ms \\
                      & MPNN            & 0.6922          & \ul{0.7780}    & 0.7768          & 0.5510          & 11,127,906 & 11.1ms \\
                      & JODIE           & 0.6922          & \ul{0.7780}    & 0.7768          & 0.5510          & 11,127,906 & 11.2ms \\
                      & EvolveGCN       & 0.7849          & 0.6593          & 0.7230          & 0.5056          & 11,127,906 & 13.9ms \\
                      & TGAT            & 0.7790          & 0.6620          & 0.7270          & 0.4900          & 11,127,906 & 11.0ms \\
\textbf{}             & SpikeNet        & 0.6940          & 0.6980          & 0.7130          & 0.4780          & 3,006,818  & 5.3ms  \\ 

                      & Diff           & \textcolor{mygreen}{+0.0136} & \textcolor{mygreen}{+0.0186} & \textcolor{mygreen}{+0.0316} & \textcolor{mygreen}{+0.0174} & -- & -- \\

\midrule
                      & \cellcolor{lightblue}\textbf{DyCAF-Net} & \cellcolor{lightblue}0.5251          & \cellcolor{lightblue}\textbf{0.2522} & \cellcolor{lightblue}\textbf{0.2509} & \cellcolor{lightblue}\textbf{0.1530} & \cellcolor{lightblue}11,127,906 & \cellcolor{lightblue}6.3ms  \\
                      & DeepWalk        & \ul{0.6806}    & 0.1822          & 0.2061          & 0.1280          & 11,127,906 & 6.4ms  \\
                      & Node2Vec        & 0.5251          & \textbf{0.2522} & \textbf{0.2509} & \textbf{0.1530} & 11,127,906 & 6.7ms  \\
                      & $\text{M}^2\text{DNE}$           & 0.5458          & 0.2356          & \ul{0.2473}    & \ul{0.1527}    & 11,127,906 & 7.0ms  \\
\textbf{City Scapes}         & DyTriad         & 0.5458          & 0.2356          & \ul{0.2473}    & \ul{0.1527}    & 11,127,906 & 6.9ms  \\
                      & MPNN            & 0.5251          & \textbf{0.2522} & \textbf{0.2509} & \textbf{0.1530} & 11,127,906 & 7.0ms  \\
                      & JODIE           & 0.5251          & \textbf{0.2522} & \textbf{0.2509} & \textbf{0.1530} & 11,127,906 & 6.5ms  \\
                      & EvolveGCN       & 0.5251          & \ul{0.2400}    & 0.2299          & 0.1446          & 11,127,906 & 6.6ms  \\
                      & TGAT            & \textbf{0.7620} & 0.1690          & 0.2050          & 0.1260          & 11,127,906 & 6.4ms  \\
\textbf{}             & SpikeNet        & 0.6080          & 0.1670          & 0.1830          & 0.1110          & 3,006,818  & 3.7ms  \\ 

                    & Diff           & \textcolor{myred}{-0.2369} & 0.0000 & 0.0000 & 0.0000 & -- & -- \\

\midrule
                      & \cellcolor{lightblue}\textbf{DyCAF-Net} & \cellcolor{lightblue}0.7538          & \cellcolor{lightblue}0.7788          & \cellcolor{lightblue}0.8176          & \cellcolor{lightblue}0.5466          & \cellcolor{lightblue}11,128,680 & \cellcolor{lightblue}13.5ms \\
                      & DeepWalk        & 0.6824          & 0.7508          & 0.7683          & 0.4920          & 11,128,680 & 12.2ms \\
                      & Node2Vec        & 0.7577          & 0.7821          & 0.8178          & 0.5482          & 11,128,680 & 11.2ms \\
                      & $\text{M}^2\text{DNE}$           & \textbf{0.7748} & \textbf{0.7893} & \textbf{0.8319} & \textbf{0.5597} & 11,128,680 & 10.9ms \\
\textbf{Traffic Density}           & DyTriad         & \ul{0.7599}    & 0.7825          & \ul{0.8209}    & \ul{0.5521}    & 11,128,680 & 12.5ms \\
                      & MPNN            & 0.7577          & \ul{0.7821}    & 0.8178          & 0.5482          & 11,128,680 & 11.2ms \\
                      & JODIE           & 0.7577          & \ul{0.7821}    & 0.8178          & 0.5482          & 11,128,680 & 11.1ms \\
                      & EvolveGCN       & 0.7077          & 0.7509          & 0.7808          & 0.5016          & 11,128,680 & 11.2ms \\
                      & TGAT            & 0.6960          & 0.7440          & 0.7710          & 0.4940          & 11,128,680 & 11.1ms \\
\textbf{}             & SpikeNet        & 0.6510          & 0.7310          & 0.7430          & 0.4610          & 3,007,208  & 5.5ms  \\    

                    & Diff           & \textcolor{myred}{-0.0210} & \textcolor{myred}{-0.0105} & \textcolor{myred}{-0.0143} & \textcolor{myred}{-0.0131} & -- & -- \\

\midrule
                      & \cellcolor{lightblue}\textbf{DyCAF-Net} & \cellcolor{lightblue}\textbf{0.8541} & \cellcolor{lightblue}\textbf{0.8429} & \cellcolor{lightblue}\textbf{0.8970} & \cellcolor{lightblue}\textbf{0.7305} & \cellcolor{lightblue}11,129,454 & \cellcolor{lightblue}13.5ms \\
                      & DeepWalk        & 0.5040          & 0.4525          & 0.4539          & 0.2811          & 11,129,454 & 12.3ms \\
                      & Node2Vec        & \ul{0.7900}    & \ul{0.7472}    & \ul{0.8426}    & \ul{0.6319}    & 11,129,454 & 10.2ms \\
                      & $\text{M}^2\text{DNE}$           & \ul{0.7900}    & \ul{0.7472}    & \ul{0.8426}    & \ul{0.6319}    & 11,129,454 & 10.2ms \\
\textbf{Animals}      & DyTriad         & \ul{0.7900}    & \ul{0.7472}    & \ul{0.8426}    & \ul{0.6319}    & 11,129,454 & 10.2ms \\
                      & MPNN            & \ul{0.7900}    & \ul{0.7472}    & \ul{0.8426}    & \ul{0.6319}    & 11,129,454 & 10.2ms \\
                      & JODIE           & \ul{0.7900}    & \ul{0.7472}    & \ul{0.8426}    & \ul{0.6319}    & 11,129,454 & 10.3ms \\
                      & EvolveGCN       & 0.1128          & 0.3523          & 0.1507          & 0.0757          & 11,129,454 & 10.2ms \\
                      & TGAT            & 0.5040          & 0.4530          & 0.4540          & 0.2810          & 11,129,454 & 10.3ms \\
           & SpikeNet        & 0.3510          & 0.4110          & 0.3470          & 0.2070          & 3,007,598  & 7.9ms  \\    

                        & Diff           & \textcolor{mygreen}{+0.0641} & \textcolor{mygreen}{+0.0957} & \textcolor{mygreen}{+0.0544} & \textcolor{mygreen}{+0.0986} & -- & -- \\

\midrule
                      & \cellcolor{lightblue}\textbf{DyCAF-Net} & \cellcolor{lightblue}\textbf{0.8545} & \cellcolor{lightblue}\textbf{0.8583} & \cellcolor{lightblue}\textbf{0.9023} & \cellcolor{lightblue}\textbf{0.5195} & \cellcolor{lightblue}11,127,519 & \cellcolor{lightblue}11.9ms \\
                      & DeepWalk        & 0.7680          & 0.7445          & 0.7704          & 0.4429          & 11,127,519 & 13.8ms \\
                      & Node2Vec        & 0.8185          & \ul{0.8465}    & \ul{0.8881}    & \ul{0.5143}    &  11,127,519          &    11.1ms    \\
                      & $\text{M}^2\text{DNE}$           & 0.8185          & \ul{0.8465}    & \ul{0.8881}    & \ul{0.5143}    & 11,127,519 & 11.1ms \\
\textbf{Construction Safety} & DyTriad         & 0.8185          & \ul{0.8465}    & \ul{0.8881}    & \ul{0.5143}    & 11,127,519 & 11.6ms \\
                      & MPNN            & 0.8185          & \ul{0.8465}    & \ul{0.8881}    & \ul{0.5143}    & 11,127,519 & 12.4ms \\
                      & JODIE           & 0.8185          & \ul{0.8465}    & \ul{0.8881}    & \ul{0.5143}    & 11,127,519 & 10.8ms \\
                      & EvolveGCN       & 0.7216          & 0.7917          & 0.8167          & 0.4541          & 11,127,519 & 10.7ms \\
                      & TGAT            & 0.7680          & 0.7450          & 0.7700          & 0.4430          & 11,127,519 & 11.7ms \\
         & SpikeNet        & \ul{0.8490}    & 0.6730          & 0.7690          & 0.4360          & 3,006,623  & 8.6ms  \\   

                    & Diff           & \textcolor{mygreen}{+0.0055} & \textcolor{mygreen}{+0.0118} & \textcolor{mygreen}{+0.0142} & \textcolor{mygreen}{+0.0052} & -- & -- \\

\midrule

& \cellcolor{lightblue}\textbf{DyCAF-Net} & \cellcolor{lightblue}\textbf{0.8582} & \cellcolor{lightblue}\textbf{0.8299} & \cellcolor{lightblue}\textbf{0.9266} & \cellcolor{lightblue}\textbf{0.6424} & \cellcolor{lightblue}11,126,358 & \cellcolor{lightblue}13.0ms \\
                      & DeepWalk        & 0.0015          & 0.0633          & 0.0008          & 0.0003          & 11,126,358 & 12.4ms \\
                      & Node2Vec        & \ul{0.7800}    & \ul{0.8183}    & \ul{0.8938}    & \ul{0.6086}    & 11,126,358 & 14.7ms \\
                      & $\text{M}^2\text{DNE}$           & \ul{0.7800}    & \ul{0.8183}    & \ul{0.8938}    & \ul{0.6086}    & 11,126,358 & 13.3ms \\
\textbf{Mask Wearing} & DyTriad        & \ul{0.7800}       & \ul{0.8183}    & \ul{0.8938}    & \ul{0.6086}       & 11,126,358         & 14.5ms                    \\
                      & MPNN            & \ul{0.7800}    & \ul{0.8183}    & \ul{0.8938}    & \ul{0.6086}    & 11,126,358 & 13.3ms \\
                      & JODIE           & \ul{0.7800}    & \ul{0.8183}    & \ul{0.8938}    & \ul{0.6086}    & 11,126,358 & 14.1ms \\
                      & EvolveGCN       & 0.1121          & 0.3341          & 0.0930          & 0.0458          & 11,126,358 & 13.3ms \\
                      & TGAT            & 0.4180          & 0.5230          & 0.4410          & 0.2150          & 11,126,358 & 13.1ms \\
       & SpikeNet        & 0.0011          & 0.0458          & 0.0006          & 0.0002          & 3,006,038  & 7.3ms  \\  

                    & Diff           & \textcolor{mygreen}{+0.0782} & \textcolor{mygreen}{+0.0116} & \textcolor{mygreen}{+0.0328} & \textcolor{mygreen}{+0.0338} & -- & -- \\

\midrule

                      & \cellcolor{lightblue}\textbf{DyCAF-Net} & \cellcolor{lightblue}\textbf{0.9999} & \cellcolor{lightblue}\textbf{1.0000} & \cellcolor{lightblue}\textbf{0.9950} & \cellcolor{lightblue}\textbf{0.9263} & \cellcolor{lightblue}11,126,358 & \cellcolor{lightblue}38.2ms \\
                      & DeepWalk        & \ul{0.9962}    & 0.9976          & \textbf{0.9950} & 0.9009          & 11,126,358 & 7.9ms  \\
                      & Node2Vec        & \textbf{0.9999} & \textbf{1.0000} & \textbf{0.9950} & \ul{0.9207}    & 11,126,358 & 8.0ms  \\
                      & $\text{M}^2\text{DNE}$           & \textbf{0.9999} & \textbf{1.0000} & \textbf{0.9950} & \ul{0.9207}    & 11,126,358 & 8.0ms  \\
\textbf{Peanuts}       & DyTriad        & \textbf{0.9999}    & \textbf{1.0000} & \textbf{0.9950} & \ul{0.9207}       & 11,126,358         & 7.9ms                     \\
                      & MPNN            & \textbf{0.9999} & \textbf{1.0000} & \textbf{0.9950} & \ul{0.9207}    & 11,126,358 & 7.8ms  \\
                      & JODIE           & \textbf{0.9999} & \textbf{1.0000} & \textbf{0.9950} & \ul{0.9207}    & 11,126,358 & 7.8ms  \\
                      & EvolveGCN       & 0.9916          & \ul{0.9982}    & \ul{0.9929}    & 0.8948          & 11,126,358 & 7.9ms  \\
                      & TGAT            & 0.9960          & 0.9980          & \textbf{0.9950} & 0.9010          & 11,126,358 & 8.1ms  \\
       & SpikeNet        & 0.9530          & 0.9730          & 0.9900          & 0.8570          & 3,006,038  & 6.6ms  \\ 

                    & Diff           & 0.0000 & 0.0000 & 0.0000 & \textcolor{mygreen}{+0.0056} & -- & -- \\

\midrule
                      & \cellcolor{lightblue}\textbf{DyCAF-Net} & \cellcolor{lightblue}\ul{0.9377}    & \cellcolor{lightblue}\textbf{0.9295} & \cellcolor{lightblue}\ul{0.9676}    & \cellcolor{lightblue}\textbf{0.8200} & \cellcolor{lightblue}11,133,711 & \cellcolor{lightblue}12.6ms \\
                      & DeepWalk        & 0.9270          & 0.9090          & 0.9494          & 0.7759          & 11,133,711 & 12.9ms \\
                      & Node2Vec        & \textbf{0.9505} & \ul{0.9219}    & \textbf{0.9696} & \ul{0.8132}    & 11,133,711 & 12.6ms \\
                      & $\text{M}^2\text{DNE}$           & \textbf{0.9505} & \ul{0.9219}    & \textbf{0.9696} & \ul{0.8132}    & 11,133,711 & 12.6ms \\
\textbf{Road Signs}   & DyTriad        & \textbf{0.9505}    & \ul{0.9219}    & \textbf{0.9696} & \ul{0.8132}       & 11,133,711         & 12.7ms                    \\
                      & MPNN            & \textbf{0.9505} & \ul{0.9219}    & \textbf{0.9696} & \ul{0.8132}    & 11,133,711 & 12.8ms \\
                      & JODIE           & \textbf{0.9505} & \ul{0.9219}    & \textbf{0.9696} & \ul{0.8132}    & 11,133,711 & 12.6ms \\
                      & EvolveGCN       & 0.9112          & 0.8960          & 0.9386          & 0.7796          & 11,133,711 & 12.6ms \\
                      & TGAT            & 0.9270          & 0.9090          & 0.9490          & 0.7760          & 11,133,711 & 12.6ms \\
         & SpikeNet        & 0.8180          & 0.8540          & 0.8880          & 0.7140          & 3,009,743  & 7.6ms  \\    

                    & Diff           & \textcolor{myred}{-0.0128} & \textcolor{mygreen}{+0.0076} & \textcolor{myred}{-0.0020} & \textcolor{mygreen}{+0.0068} & -- & -- \\

\midrule
                      & \cellcolor{lightblue}\textbf{DyCAF-Net} & \cellcolor{lightblue}\ul{0.6936}    & \cellcolor{lightblue}\ul{0.7737}    & \cellcolor{lightblue}\ul{0.6943}    & \cellcolor{lightblue}\textbf{0.4575} & \cellcolor{lightblue}11,128,680 & \cellcolor{lightblue}13.6ms \\
                      & DeepWalk        & 0.5795          & 0.6657          & 0.5991          & 0.3326          & 11,128,680 & 14.1ms \\
                      & Node2Vec        & \textbf{0.7012} & \textbf{0.7944} & \textbf{0.6988} & \ul{0.4429}    & 11,128,680 & 12.3ms \\
                      & $\text{M}^2\text{DNE}$           & \textbf{0.7012} & \textbf{0.7944} & \textbf{0.6988} & \ul{0.4429}    & 11,128,680 & 11.6ms \\
\textbf{Street Work}  & DyTriad        & \textbf{0.7012}    & \textbf{0.7944} & \textbf{0.6988} & \ul{0.4429}       & 11,128,680         & 11.7ms                    \\
                      & MPNN            & \textbf{0.7012} & \textbf{0.7944} & \textbf{0.6988} & \ul{0.4429}    & 11,128,680 & 13.4ms \\
                      & JODIE           & \textbf{0.7012} & \textbf{0.7944} & \textbf{0.6988} & \ul{0.4429}    & 11,128,680 & 12.0ms \\
                      & EvolveGCN       & 0.5093          & 0.4546          & 0.4251          & 0.2212          & 11,128,680 & 11.8ms \\
                      & TGAT            & 0.5800          & 0.6660          & 0.5990          & 0.3330          & 11,128,680 & 11.7ms \\
         & SpikeNet        & 0.6520          & 0.5180          & 0.5070          & 0.2840          & 3,007,208  & 8.7ms  \\    

                    & Diff           & \textcolor{myred}{-0.0076} & \textcolor{myred}{-0.0207} & \textcolor{myred}{-0.0045} & \textcolor{mygreen}{+0.0146} & -- & -- \\

\midrule
                      & \cellcolor{lightblue}\textbf{DyCAF-Net} & \cellcolor{lightblue}\textbf{0.7557} & \cellcolor{lightblue}\textbf{0.6087} & \cellcolor{lightblue}\ul{0.6435}    & \cellcolor{lightblue}\ul{0.4557}    & \cellcolor{lightblue}11,130,228 & \cellcolor{lightblue}12.0ms \\
                      & DeepWalk        & 0.6855          & 0.4762          & 0.4899          & 0.3427          & 11,130,228 & 12.7ms \\
                      & Node2Vec        & \ul{0.7552}    & \ul{0.6060}    & \textbf{0.6466} & \textbf{0.4630} & 11,130,228 & 11.4ms \\
                      & $\text{M}^2\text{DNE}$           & \ul{0.7552}    & \ul{0.6060}    & \textbf{0.6466} & \textbf{0.4630} & 11,130,228 & 11.4ms \\
\textbf{Wine Labels}  & DyTriad        & \ul{0.7552}       & \ul{0.6060}    & \textbf{0.6466} & \textbf{0.4630}    & 11,130,228         & 11.4ms                    \\
                      & MPNN            & \ul{0.7552}    & \ul{0.6060}    & \textbf{0.6466} & \textbf{0.4630} & 11,130,228 & 11.5ms \\
                      & JODIE           & \ul{0.7552}    & \ul{0.6060}    & \textbf{0.6466} & \textbf{0.4630} & 11,130,228 & 11.5ms \\
                      & EvolveGCN       & 0.6886          & 0.5482          & 0.5700          & 0.4039          & 11,130,228 & 11.7ms \\
                      & TGAT            & 0.6860          & 0.4760          & 0.4900          & 0.3430          & 11,130,228 & 11.5ms \\
        & SpikeNet        & 0.6290          & 0.4380          & 0.4310          & 0.3040          & 3,007,988  & 7.0ms  \\    

                    & Diff           & \textcolor{mygreen}{+0.0005} & \textcolor{mygreen}{+0.0027} & \textcolor{myred}{-0.0031} & \textcolor{myred}{-0.0073} & -- & -- \\

\midrule
                      & \cellcolor{lightblue}\textbf{DyCAF-Net} & \cellcolor{lightblue}\textbf{0.8090} & \cellcolor{lightblue}0.5841          & \cellcolor{lightblue}\textbf{0.6749} & \cellcolor{lightblue}\textbf{0.4815} & \cellcolor{lightblue}11,126,358 & \cellcolor{lightblue}16.5ms \\
                      & DeepWalk        & 0.5517          & 0.7202          & 0.6016          & \ul{0.3824}    & 11,126,358 & 13.3ms \\
                      & Node2Vec        & 0.4349          & \textbf{0.8192} & 0.4934          & 0.3405          & 11,126,358 & 13.3ms \\
                      & $\text{M}^2\text{DNE}$           & 0.4349          & \textbf{0.8192} & 0.4934          & 0.3405          & 11,126,358 & 13.5ms \\
\textbf{Axial MRI}        & DyTriad         & 0.4349          & \textbf{0.8192} & 0.4934          & 0.3405          & 11,126,358 & 13.3ms \\
                      & MPNN            & 0.4349          & \textbf{0.8192} & 0.4934          & 0.3405          & 11,126,358 & 13.4ms \\
                      & JODIE           & 0.4349          & \textbf{0.8192} & 0.4934          & 0.3405          & 11,126,358 & 13.4ms \\
                      & EvolveGCN       & 0.4792          & 0.7113          & 0.5082          & 0.3237          & 11,126,358 & 16.4ms \\
                      & TGAT            & \ul{0.5520}    & 0.7200          & \ul{0.6020}    & 0.3820          & 11,126,358 & 13.3ms \\
        & SpikeNet        & 0.3770          & \ul{0.7360}    & 0.5530          & 0.3120          & 3,006,038  & 14.1ms \\    

                    & Diff           & \textcolor{mygreen}{+0.2570} & \textcolor{myred}{-0.2351} & \textcolor{mygreen}{+0.0729} & \textcolor{mygreen}{+0.0991} & -- & -- \\
                    
\midrule
                      & \cellcolor{lightblue}\textbf{DyCAF-Net} & \cellcolor{lightblue}\textbf{0.9706} & \cellcolor{lightblue}\textbf{0.9891} & \cellcolor{lightblue}\textbf{0.9941} & \cellcolor{lightblue}\textbf{0.8691} & \cellcolor{lightblue}11,126,745 & \cellcolor{lightblue}14.3ms \\
                      & DeepWalk        & 0.8562          & 0.8938          & 0.9524          & 0.7542          & 11,126,745 & 15.0ms \\
                      & Node2Vec        & \ul{0.8942}    & \ul{0.9379}    & \ul{0.9648}    & \ul{0.8112}    & 11,126,745 & 14.4ms \\
                      & $\text{M}^2\text{DNE}$           & \ul{0.8942}    & \ul{0.9379}    & \ul{0.9648}    & \ul{0.8112}    & 11,126,745 & 12.8ms \\
\textbf{Furniture}    & DyTriad         & \ul{0.8942}    & \ul{0.9379}    & \ul{0.9648}    & \ul{0.8112}    & 11,126,745 & 13.4ms \\
                      & MPNN            & \ul{0.8942}    & \ul{0.9379}    & \ul{0.9648}    & \ul{0.8112}    & 11,126,745 & 13.0ms \\
                      & JODIE           & \ul{0.8942}    & \ul{0.9379}    & \ul{0.9648}    & \ul{0.8112}    & 11,126,745 & 13.0ms \\
                      & EvolveGCN       & 0.5683          & 0.5072          & 0.6796          & 0.4433          & 11,126,745 & 13.3ms \\
                      & TGAT            & 0.8560          & 0.8940          & 0.9520          & 0.7540          & 11,126,745 & 14.5ms \\
         & SpikeNet        & 0.7390          & 0.9300          & 0.8830          & 0.6850          & 3,006,233  & 9.4ms  \\    

                        & Diff           & \textcolor{mygreen}{+0.0764} & \textcolor{mygreen}{+0.0512} & \textcolor{mygreen}{+0.0293} & \textcolor{mygreen}{+0.0579} & -- & -- \\

\midrule                      
                      & \cellcolor{lightblue}\textbf{DyCAF-Net} & \cellcolor{lightblue}\ul{0.9105}    & \cellcolor{lightblue}\textbf{0.8717} & \cellcolor{lightblue}\ul{0.8593}    & \cellcolor{lightblue}\textbf{0.5774} & \cellcolor{lightblue}11,130,228 & \cellcolor{lightblue}13.6ms \\
                      & DeepWalk        & 0.6633          & 0.5874          & 0.6603          & 0.3776          & 11,130,228 & 46.5ms \\
                      & Node2Vec        & \textbf{0.9245} & \ul{0.8616}    & \textbf{0.8682} & \ul{0.5770}    & 11,130,228 & 15.3ms \\
                      & $\text{M}^2\text{DNE}$           & \textbf{0.9245} & \ul{0.8616}    & \textbf{0.8682} & \ul{0.5770}    & 11,130,228 & 14.0ms \\
\textbf{X-Ray}        & DyTriad         & \textbf{0.9245} & \ul{0.8616}    & \textbf{0.8682} & \ul{0.5770}    & 11,130,228 & 13.9ms \\
                      & MPNN            & \textbf{0.9245} & \ul{0.8616}    & \textbf{0.8682} & \ul{0.5770}    & 11,130,228 & 15.3ms \\
                      & JODIE           & \textbf{0.9245} & \ul{0.8616}    & \textbf{0.8682} & \ul{0.5770}    & 11,130,228 & 15.3ms \\
                      & EvolveGCN       & 0.2713          & 0.0786          & 0.0294          & 0.0085          & 11,130,228 & 15.5ms \\
                      & TGAT            & 0.6630          & 0.5870          & 0.6600          & 0.3780          & 11,130,228 & 21.2ms \\
          & SpikeNet        & 0.5610          & 0.5680          & 0.5540          & 0.2950          & 3,007,988  & 16.7ms   \\

                     & Diff           & \textcolor{myred}{-0.0140} & \textcolor{mygreen}{+0.0101} & \textcolor{myred}{-0.0089} & \textcolor{mygreen}{+0.0004} & -- & -- \\

\bottomrule
\end{longtable}
\end{sc}
}
\normalsize

\clearpage
\twocolumn

\section{Limitations}
While DyCAF-Net improves detection for rare and occluded objects, it can slightly reduce precision in some cases due to aggressive recalibration. Gains are also less pronounced in low-imbalance datasets. Finally, our study focuses on static image detection; future work will extend DyCAF-Net to video and few-shot learning settings.

% \let \clearpage
% \newpage
% \vspace*{330px}

% or
% \appendix[Related Works]  % for no appendix heading
% \input{sec/2_rwork}


% ============================================
% \appendices
%Appendix one text goes here %\cite{Roberg2010}.

%\section{}

%\section*{Acknowledgment}

%The authors would like to thank D. Root for the loan of the SWAP. The SWAP that can ONLY be usefull in Boulder...

\ifCLASSOPTIONcaptionsoff
  \newpage
\fi

% The ``triggered" command can be changed if desired:


%\begin{thebibliography}{1}


{
\tiny
\bibliographystyle{IEEEtran}
\bibliography{IEEEabrv,main}
}

\vfill

\appendix[Related Works]  % for no appendix heading
\section{Related Work}
\label{appendix:related_works}

\subsection{Object Detection Architectures}
Modern object detectors such as YOLOv8 \cite{terven_comprehensive_2023} and Faster R-CNN \cite{ren_faster_2015} rely on three core components: a backbone for hierarchical feature extraction, a neck for multi-scale fusion, and a head for bounding box regression and classification. Among these, the neck architecture plays a critical role in addressing scale variance, a persistent challenge in detection tasks. Early works like Feature Pyramid Networks (FPN) \cite{lin_feature_2017} introduced top-down pathways to merge semantically rich deep-layer features with spatially precise shallow-layer ones. Subsequent advancements, such as PANet \cite{liu_path_2018}, augmented this design with bottom-up aggregation to improve localization accuracy, while BiFPN \cite{tan_efficientdet_2020} optimized cross-scale connections using learnable weights. While effective, these methods depend on predefined fusion heuristics (e.g., fixed upsampling or summation rules), limiting their adaptability to dynamic scenes where scale relationships vary significantly across inputs. This rigidity motivates our work to introduce input-conditioned fusion in DyCAF-Net, enabling scene-aware feature aggregation.

\subsection{Attention Mechanisms in Vision}
Attention mechanisms have emerged as powerful tools for refining discriminative features. SENet \cite{hu_squeeze-and-excitation_2018} pioneered channel-wise recalibration, adaptively emphasizing informative feature maps. Building on this, CBAM \cite{woo_cbam_2018} integrated spatial attention to highlight salient regions. Dynamic architectures like DyNet \cite{han_dynamic_2021} generalized these principles by conditioning attention parameters on input features, enabling context-aware adaptation. Recent innovations such as Deformable Attention Transformers (DAT) \cite{zhu_deformable_2021} further leverage deformable sampling to focus on task-relevant areas. However, these frameworks operate in a class-agnostic manner, neglecting nuanced feature interactions required to disentangle occluded objects or mitigate class imbalance in long-tailed datasets. DyCAF-Net addresses this gap through class-aware attention, dynamically modulating spatial and channel features based on object categories.

\subsection{Multi-Scale Fusion with Implicit Modeling}
Implicit models like Deep Equilibrium Networks (DEQ) \cite{bai_deep_2019} and their multiscale variants \cite{bai_multiscale_2020} offer memory-efficient alternatives to traditional deep networks by solving for equilibrium states instead of storing intermediate activations. In object detection, Recursive-FPN \cite{wang_arfp_2022} adopted recursive connections to iteratively refine features, implicitly modeling equilibrium dynamics. While these methods reduce memory overhead, they lack explicit propagation of class-specific contextual cues across scales, limiting their ability to resolve ambiguities in cluttered scenes. DyCAF-Net bridges this gap by integrating implicit equilibrium fusion with class-aware feature modulation, enabling memory-efficient yet semantically rich multi-scale reasoning.

\subsection{Class-Aware Feature Adaptation}
Addressing class imbalance, prior works have primarily focused on loss function modifications. Class-Balanced Loss \cite{cui_class-balanced_2019} and Equalization Loss \cite{li_adaptive_2022} adjust training objectives to counteract bias toward frequent classes, while RepMet \cite{karlinsky_repmet_2019} employs metric learning to isolate class-specific embeddings. Architecturally, DyHead \cite{dai_dynamic_2021} unified scale, spatial, and task-aware attention but omitted explicit class-wise feature modulation, leaving a gap in leveraging semantic hierarchies for detection.

Conventional neck architectures, including PANet \cite{liu_path_2018} and BiFPN \cite{tan_efficientdet_2020}, exhibit three critical shortcomings. First, their static fusion rules fail to adapt to input-dependent scale relationships, leading to suboptimal feature aggregation in heterogeneous scenes. Second, their attention mechanisms lack dynamic conditioning on spatial and channel dimensions, limiting robustness to occlusions or background clutter. Third, class-agnostic feature aggregation overlooks semantic distinctions between object categories, hindering performance in long-tailed or fine-grained datasets. DyCAF-Net bridges these gaps through dynamic dual attention mechanisms that adaptively recalibrate spatial and channel features, implicit multi-scale equilibrium modeling for memory-efficient fusion, and class-aware feature weighting in the detection head to prioritize discriminative cues across object categories.

%\end{thebibliography}

% \vfill


\end{document}